\pdfoutput=1
\documentclass[11pt]{article}

\usepackage[]{acl}
\usepackage{times}
\usepackage{latexsym}

\usepackage[T1]{fontenc}
\usepackage[utf8]{inputenc}

\usepackage{microtype}
\usepackage{graphicx}
\usepackage{algorithm}
\usepackage{algorithmic}
\usepackage{enumerate}
\usepackage{subfigure}
\usepackage{amsmath}
\usepackage{tabularray}
\usepackage{graphicx}
\usepackage{makecell}
\usepackage{multirow}
\usepackage{booktabs}
\usepackage{inconsolata}
\usepackage{xcolor} 
\usepackage{hyperref}
\title{A Rationale-centric Counterfactual Data Augmentation Method for Cross-Document Event Coreference Resolution}

\author{
    Bowen Ding\textsuperscript{\rm 2, \footnotemark[1], \footnotemark[3]},
    Qingkai Min\textsuperscript{\rm 1,2,\footnotemark[1]},  
    Shengkun Ma\textsuperscript{\rm 4, \footnotemark[3]}, Yingjie Li\textsuperscript{\rm 2}, Linyi Yang\textsuperscript{\rm 2,3, \footnotemark[2]} and Yue Zhang\textsuperscript{\rm 2,3,\footnotemark[2]}
    \\
    \textsuperscript{1} Zhejiang University \\
    \textsuperscript{2} School of Engineering, Westlake University \\
    \textsuperscript{3} Institute of Advanced Technology, Westlake Institute for Advanced Study \\
    \textsuperscript{4} Beijing University of Posts and Telecommunications\\
    $^{2}$\texttt{\{dingbowen, minqingkai, liyingjie, yanglinyi, zhangyue\}@westlake.edu.cn} \\
    $^{4}$\texttt{mashengkun@bupt.edu.cn} \\
}

\begin{document}
\maketitle
\renewcommand{\thefootnote}{\fnsymbol{footnote}}
\footnotetext[1]{Contributed equally.}
\footnotetext[2]{Corresponding authors.}
\footnotetext[3]{Work done at Westlake University as an intern.}
\renewcommand{\thefootnote}{\arabic{footnote}}

\begin{abstract}
\label{sec:abstract}
Based on Pre-trained Language Models (PLMs), event coreference resolution (ECR) systems have demonstrated outstanding performance in clustering coreferential events across documents. However, the state-of-the-art system exhibits an excessive reliance on the `triggers lexical matching' spurious pattern in the input mention pair text. We formalize the decision-making process of the baseline ECR system using a Structural Causal Model (SCM), aiming to identify spurious and causal associations (i.e., rationales) within the ECR task. Leveraging the debiasing capability of counterfactual data augmentation, we develop a rationale-centric counterfactual data augmentation method with LLM-in-the-loop. This method is specialized for pairwise input in the ECR system, where we conduct direct interventions on triggers and context to mitigate the spurious association while emphasizing the causation. Our approach achieves state-of-the-art performance on three popular cross-document ECR benchmarks and demonstrates robustness in out-of-domain scenarios. 
\end{abstract}

\section{Introduction}
\label{sec:Introduction}
The goal of cross-document event coreference resolution (ECR) is to group event mentions referring to the same real-world event together across documents. It is an essential task in NLP and has provided valuable prior event-related knowledge for many downstream tasks, e.g., topic detection and tracking~\cite{Allan1998TopicDA}, multi-hop question answering~\cite{yang-etal-2018-hotpotqa} and information extraction~\cite{humphreys-etal-1997-event}. In real life, event coreference systems commonly assist decision-makers in important fields such as intelligence analysis and security event warnings~\cite{Palantir}.

\begin{figure}[t]
    \centering
    \includegraphics[width=1.0\columnwidth]{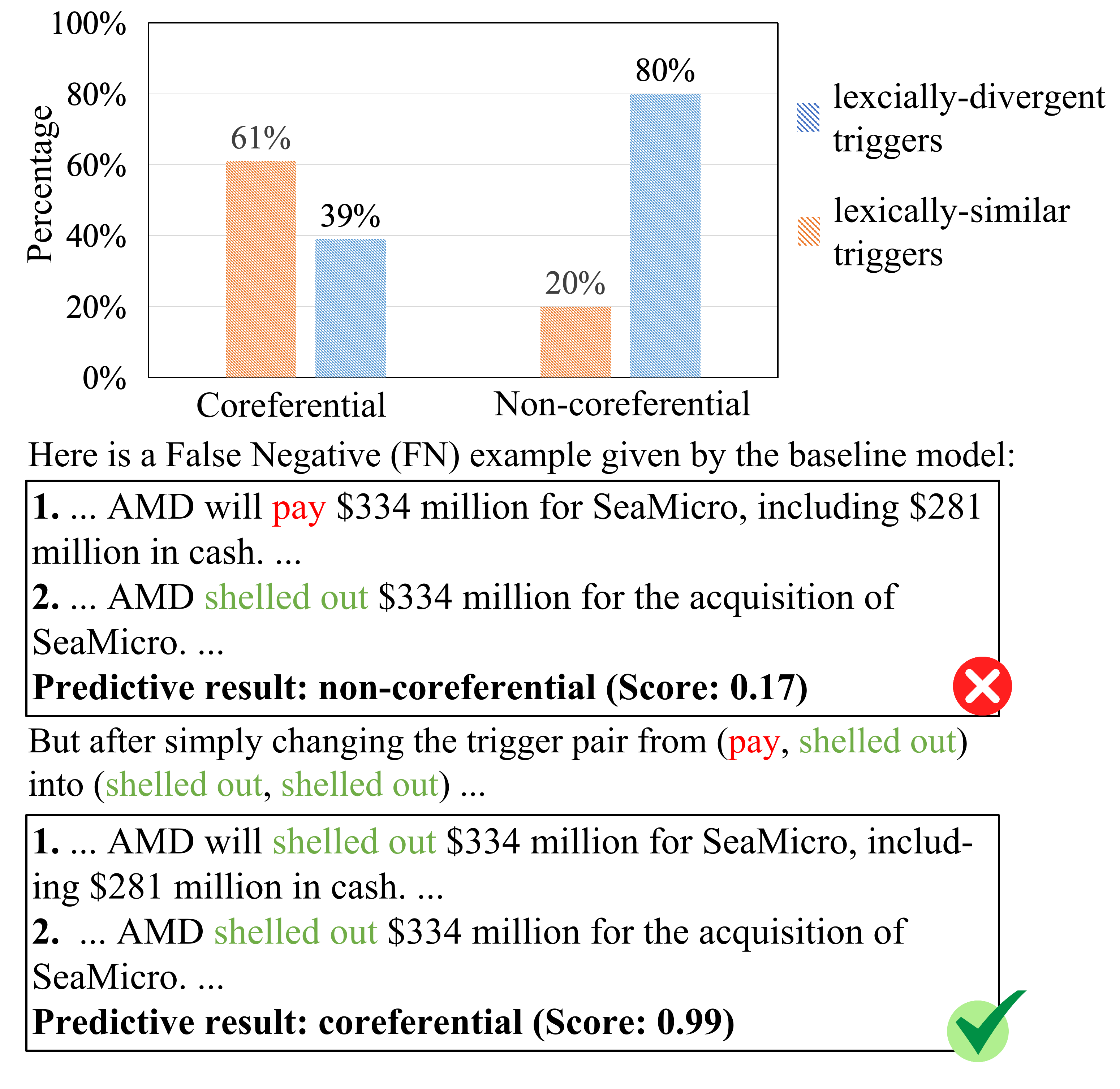} 
    \caption{The distribution of `triggers lexical matching' in mention pairs from ECB+ training set, along with a false negative example from ~\citeauthor{held-etal-2021-focus}'s system which shows that forcing the event trigger in the first mention to lexically match the second one causes a significant change in the predicted coreference score. }
    \label{fig:intro_pictures}
\end{figure}

To resolve the task, existing state-of-the-art ECR systems perform binary classification to pairwise compare event mentions~\cite{barhom2019revisiting,caciularu2021cdlm,held-etal-2021-focus,yu2023pairwise}. In their pipelines, a fraction of coreferential and non-coreferential mention pairs are retrieved from the corpus to fine-tune a cross-encoder, which is used as a coreference scorer to gauge the likelihood of pairwise events being coreferential. Finally, coreferential mentions are merged into clusters based on the predicted coreference score.   

However, most coreference scorers are troubled by the curse of `triggers lexical matching', which is also discussed in previous works~\cite{ravi2023happens,ahmed20232}. The histogram in Figure~\ref{fig:intro_pictures} demonstrates that when constructing event-mention pairs, it is natural that coreferential mentions frequently share lexically similar~\footnote{The differentiation criteria refers to Appx. Sec~\ref{sec:Triggers_lexical_similarity}.} event triggers, whereas non-coreferential mentions typically have lexically divergent ones. 

This skewed feature distribution brings a bias where lexically similar trigger words often correspond to coreference, especially for trigger-centric ECR systems~\cite{held-etal-2021-focus,yu2023pairwise} using the event representation in Appx Sec~\ref{sec:representation}. However, what truly determines the ECR outcome of event mentions is the coreference of event-relevant arguments, which include (non-)human participants, times, locations, and actions (i.e., event triggers)~\cite{cybulska2015bag}. In other words, these deeper semantic features constitute the rationales of the ECR task, as they demonstrate the task's corresponding causal associations. Unfortunately, the example in Figure~\ref{fig:intro_pictures} implies that 
the state-of-the-art system~\cite{held-etal-2021-focus} relies too much on the surface feature of trigger term similarity but ignores contextual semantics. To resolve this issue, we think about adjusting the distribution of key features in the training data through data governance. 

\begin{figure*}[t]
    \centering
    \includegraphics[width=1.0\textwidth]{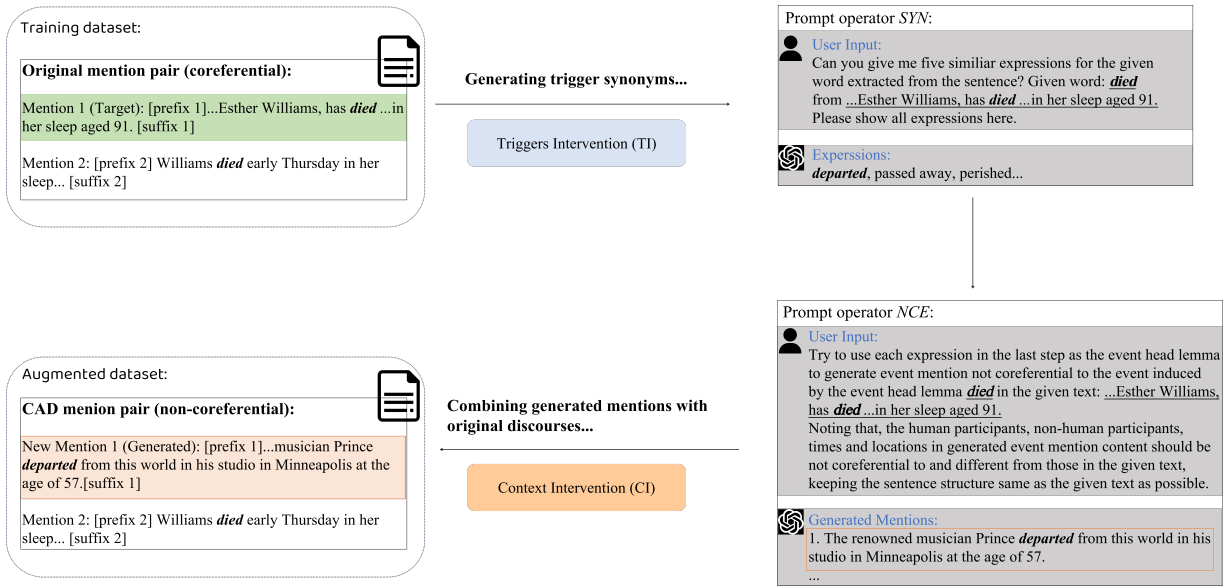} 
    \caption{The procedure of our rationale-centric counterfactual DA with LLM-in-the-loop (LLM-RCDA).}
    \label{fig:DA_process}
\end{figure*}

Counterfactual data augmentation (DA) is a promising way for debiasing the classification system~\cite{garg2019counterfactual,madaan2021generate}, which enhances the robust causal thinking ability of models with a human-like logic: `\textit{What the output label would be if certain phrases within the input text were altered?}'. In practice, we can intervene with rationales in the original example input text to ensure minimal editing to flip the output label, thus generating counterfactual augmented data (CAD). The minimal editing constraint is to prevent the introduction of unnecessary noise into the augmented data, which allows the model trained with CAD to focus directly on the causal associations from rationales, rather than on other parts of the input text~\cite{keane2020good}.

Given this, we propose LLM-RCDA, a \textbf{r}ationale-centric \textbf{c}ounterfactual \textbf{DA} method with a \textbf{l}arge \textbf{l}anguage \textbf{m}odel (LLM) in the loop, aiming to enhance the model to think causally and understand deeper semantics in the pairwise context. As shown in Figure~\ref{fig:DA_process}, our method focuses on intervening triggers and rationales in the event-mention sentence. In the phase of trigger intervention, lexically divergent synonyms of the original trigger are generated to force the system to capture the coreferential meaning between triggers, while in the phase of context intervention, we use the LLM to merely change the rationales of the target event-mention based on prompts, and keep the discourse of the event-mention unchanged. 

To evaluate the efficacy of our method, we evaluate our method on three popular cross-document ECR benchmarks: ECB+~\cite{cybulska-vossen-2014-using}, Football Coreference Corpus (FCC)~\cite{bugert-etal-2021-generalizing} and Gun Violence Corpus (GVC)~\cite{vossen-etal-2018-dont}. Our enhanced system achieves state-of-the-art performance on all of them, with improvements varying from 1.8 to 2.3 CoNLL F1 over baselines. On ECB+, our approach significantly surpasses the performance of directly employing LLMs, showcasing its superiority to the current LLM-QA paradigm in the task. Additionally, the cross-corpus experiment on the out-of-the-domain data shows a robustness improvement of our method, with a 7.2 CoNLL F1 gain over the baseline. 

To the best of our knowledge, we are the first to evaluate and analyze the performance of popular LLMs on the cross-document ECR benchmark, and the first to formalize the decision process of the mainstream ECR model from a causal view. Moreover, we are the first to utilize rationale-centric CAD generated by the LLM to causally enhance the ECR system~\footnote{code: https://github.com/Danield21/Rationale4CDECR}.

\section{Related Work}
\label{sec:related work}
\textbf{Event Coreference Resolution}
Currently, the pre-trained language models~\cite{devlin-etal-2019-bert,liu2019roberta} have significantly enhanced the contextual semantics of text data. In recent ECR systems~\cite{kenyon-dean-etal-2018-resolving,caciularu2021cdlm,held-etal-2021-focus,chen-etal-2023-cross}, the pairwise representation of events (known as cross-encoding) becomes mainstream. Such representation combines contextual embeddings with pairwise token-level trigger embeddings to represent the event-mention pair and expects the embeddings to encode the event-relevant argument information implicitly. Some other works enhanced the pairwise events representation by explicitly fusing the encoding of event-relevant arguments which are extracted by Semantic Role Labeling (SRL) systems~\cite{barhom2019revisiting,zeng2020event,yu2023pairwise}, achieving success in performance improvements.

In our approach, we also emphasize the crucial features that influence event coreference, such as argument features. However, we do not alter the model structure or existing representation methods. Instead, we induce the model to learn these key features through rationale-centric counterfactual data augmentation, thereby enhancing the causal reasoning capability of the ECR system.

\noindent \textbf{Counterfactual Data Augmentation}
Counterfactual data augmentation is widely used in NLP tasks to improve the system's performance and robustness. The methods for generating counterfactual augmented data (CAD) vary across tasks, such as Sentiment Analysis (SA)~\cite{yang2022exploring}, Natural Language Inference (NLI)~\cite{pope2021text,robeer-etal-2021-generating-realistic} and Neural Machine Translation (NMT)~\cite{liu-etal-2021-counterfactual}. In early works, CAD generation either relies on a human-in-the-loop system~\cite{kaushik2020learning,srivastava2020robustness} or relies on PLMs~\cite{tucker2021modified,wu2021polyjuice} or external knowledge bases automatically~\cite{wang2020robustness,yang2022exploring}. Recently,~\citeauthor{li2023large} explored and confirmed the feasibility of using LLMs to generate CAD on SA, NLI, NER (Named Entity Recognition) and RE (Relation Extraction) tasks, demonstrating good efficiency. 

Our work is the first one specifically designed for the ECR task, which also involves the LLM-in-the-loop to automatically and efficiently construct the required CAD that meets the task's causal requirements. 

\section{Baselines}
\label{sec:baselines}
\textbf{LLM}
Currently, little work has been done to evaluate LLMs' performance on cross-document ECR. To achieve this task, LLMs must possess the ability to comprehend and process a long context for understanding and comparing event mentions across multiple documents. Therefore, we utilize Claude-2 (100K maximum input length)~\cite{Anthropic} and GPT-4 (8K maximum input length)~\cite{OpenAI}, two LLMs with strong long context comprehensions, to perform the evaluation. We compare the zero-shot results of LLMs with a rule-based system which employs the same head lemma matching technique~\cite{barhom2019revisiting}, an end-to-end neural system~\cite{cattan2020streamlining}, the state-of-the-art pipeline system~\cite{held-etal-2021-focus} and our causally enhanced system. 

\noindent \textbf{Fully fine-tuned Baseline}
Our method is built upon the state-of-the-art ECR system~\cite{held-etal-2021-focus}, which serves as our main baseline. ~\citeauthor{held-etal-2021-focus} applies the discourse coherence theory to create event mention pairs for training and inference. For each event mention, they retrieve the \textit{K} nearest mentions in a trained event representation space to establish matches. These event-mention pairs are encoded by RoBERTa-large~\cite{liu2019roberta} ( Appx. Sec~\ref{sec:representation}), and then fed to fine-tune the coreference scorer. During inference, the system prunes non-coreferential mention pairs and merges the remaining greedily to construct the coreferential cluster.
On top of that, we also compare with an ELMo-based system~\cite{barhom2019revisiting}, where the entity and event coreference are jointly modelled; an end-to-end cross-document coreference resolution system for both event and entity~\cite{cattan2020streamlining}; a robust feature-based system~\cite{bugert-etal-2021-generalizing}; a CDLM-based system~\cite{caciularu2021cdlm}, which uses a larger longformer~\cite{beltagy2020longformer} model for document-level representation; a system with pairwise triggers and arguments representation~\cite{yu2023pairwise} and a system trained with pruned mention pairs~\cite{ahmed20232}.

\section{Method}
\label{sec:method}
\begin{figure}[t]
    \centering
    \includegraphics[width=0.9\columnwidth]{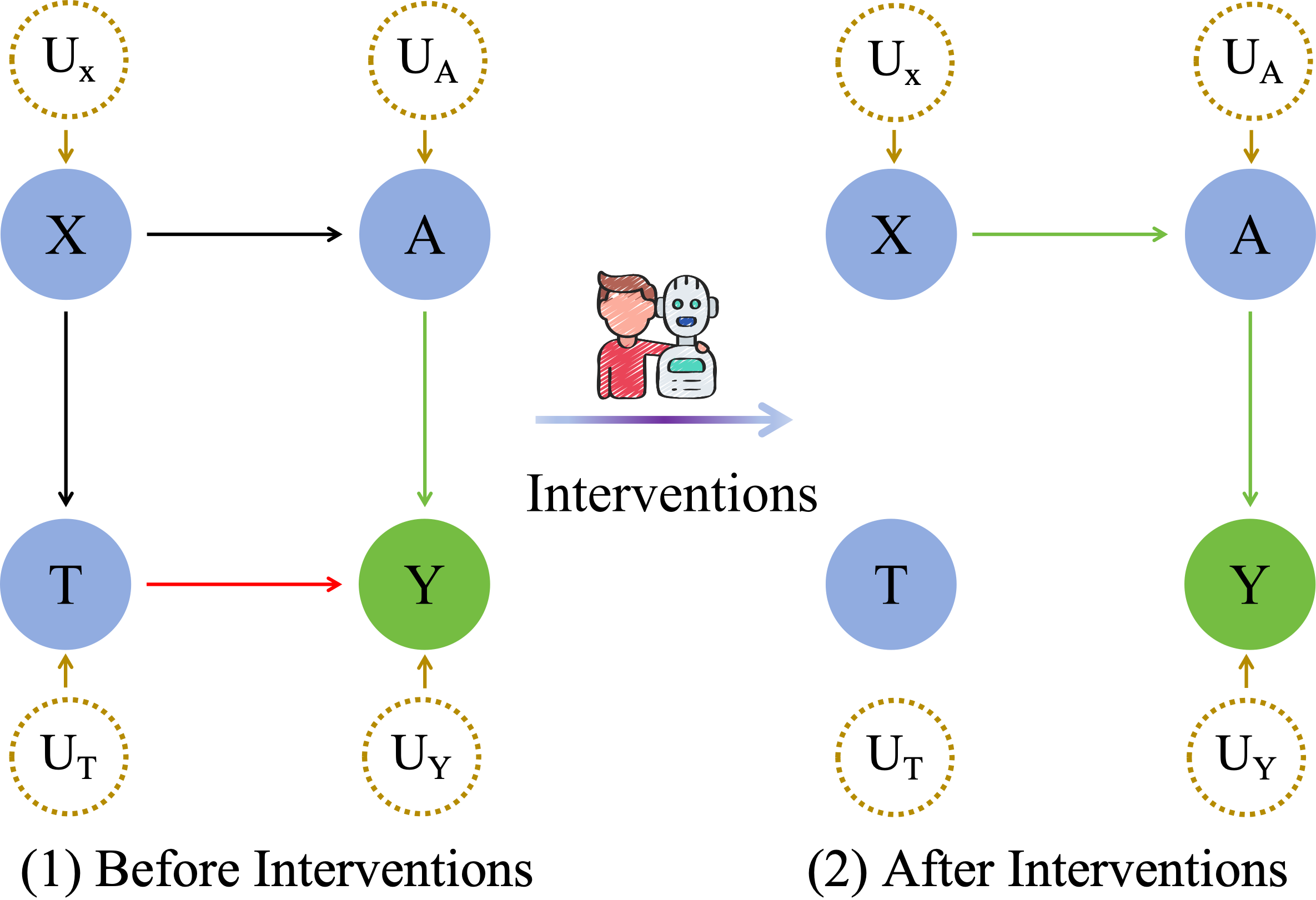} 
    \caption{SCM illustration. (1) stimulates the decision process of the baseline ECR system; (2) shows the decision process of the causally enhanced system after interventions.}
    \label{fig:causal_plot}
\end{figure}

We analyze the decision process of the ECR system on event coreference with Structural Causal Model (SCM) \cite{Pearl2000-PEACMR}. Formally, the event coreference process of the baseline ECR system into the following equation:
\begin{equation}
    \small
    \label{eq:coref_dependency}
    Y=f\left( T\left( X\right)  ,A\left( X\right)  ,U\right)
\end{equation}
where $X$ and $Y$ represent the input pairwise data and the output label, $A$ represents the semantic coreference of all counterpart event-relevant arguments which include times, locations, participants and actions (i.e., triggers) in the context of the pairwise input, $T$ denotes the scenario of triggers matching lexically and $U$ refers to the unobserved variable. Equation~\ref{eq:coref_dependency} demonstrates the fact that `triggers lexical matching' influences the prediction heavily in our baseline ECR system. At the same time, the coreferential counterparts of event-relevant arguments are rationales for ECR according to the definition of event coreference~\cite{cybulska2015bag}.  

\subsection{Causality Analysis and Interventions}
\label{sec:scm_causal_intervene}
Equation~\ref{eq:coref_dependency} can be represented by the causal graph in Figure~\ref{fig:causal_plot} (1), where the input pairwise data \textit{X} serves as a confounder of triggers matching (\textit{T}) and the coreference result (\textit{Y}), indicating a backdoor path $T\leftarrow X\rightarrow Y$, where the ECR system does not capture the causality by recognizing the semantic rationales for ECR provided in the context completely but relies on the linguistic surface feature of the trigger pair, which is a spurious association to the coreference prediction. 

To address this problem, we perform the Trigger Intervention (TI) on the path $X\rightarrow T$, as well as the Context Intervention (CI) on the path $X\rightarrow A\rightarrow Y$. TI aims to decompose the spurious association from `triggers lexical matching' and induce the ECR system to understand more about semantics. For implementation, we use prompt operator \textit{SYN} (Appx. Table~\ref{tab:coref_syn_cf_prompt}) to generate semantically related but lexically divergent synonyms of existing trigger terms, which expands their limited expressions in the corpus. This allows to adjust the distribution of trigger-matching features while keeping the meaning of the original trigger pairs. As for CI, it aims to enhance the baseline system causally by emphasizing the causal association from rationales. We decide to use counterfactual data augmentation for such intervention.

Therefore, we develop LLM-RCDA as shown in Figure~\ref{fig:DA_process}, a \textbf{r}ationale-centric \textbf{c}ounterfactual \textbf{d}ata \textbf{a}ugmentation method specifically for ECR. The method combines TI and CI, generating CAD to emphasize causal features while simultaneously not exacerbating the original skewed trigger-matching bias for the ECR system. We will introduce the algorithm in Section~\ref{sec:counterfacual_gen}. 

\subsection{Counterfactual Generation}
\label{sec:counterfacual_gen}
As demonstrated in Algorithm~\ref{alg:algorithm}, we design a LLM-in-the-loop mechanism to automatically generate \textbf{c}ounterfactual \textbf{a}ugmented \textbf{d}ata (CAD) candidates for a given original event-mention pair \textit{MP}. All prompt operators are presented in Appx. Sec~\ref{sec:prompt_for_data_gen}. Since we follow the discourse setup in the system of~\citeauthor{held-etal-2021-focus}, each event-mention text is associated with a maximum of $2w+1$ sentences totally in a discourse context. Therefore, \textit{MP} can be symbolized as (\textit{S}$^{\left( 1\right)  }_{i-w}$...\textit{S}$^{\left( 1\right)  }_{i}$...\textit{S}$^{\left( 1\right)  }_{i+w}$; \textit{S}$^{\left( 2\right)  }_{j-w}$...\textit{S}$^{\left( 2\right)  }_{j}$...\textit{S}$^{\left( 2\right)  }_{j+w}$), where \textit{S}$^{\left( 1\right)  }_{.}$ represents the sentence associated with the first event-mention in the pair, and \textit{S}$^{\left( 2\right)  }_{.}$  represents the sentence associated with the second one. For simplicity, we divide the text of an event-mention into prefix sentences (\textit{S}${}^{\left( 1\right)  }_{i-w}$,...,\textit{S}$^{\left( 1\right)  }_{i-1}$), mention sentence S$^{\left( 1\right)  }_{i}$ and suffix sentences (\textit{S}$^{\left( 1\right)  }_{i+1}$,...,\textit{S}$^{\left( 1\right)  }_{i+w}$). 

We adjust the text related to the first mention when generating CAD. Firstly, we select a mention sentence within the \textit{MP} as the \textit{target mention}, depending on the original mention pair's label (lines 2-3\&9-10). Then, \textit{target mention} undergoes `Trigger Intervention' via synonyms generation (lines 4\&11). Following that, several mention candidates are generated, each of which incorporates the trigger synonym from the last step, and is either non-coreferential or coreferential to the \textit{target mention} (lines 5\&12). These mention candidates are stored in a set \textit{S}$_{gens}$.

If the original \textit{MP} is coreferential, generating CAD becomes somewhat simpler. We only need to sequentially replace the target mention sentence $S^{(1)}_{i}$ with each of the generated candidates $s_{g}$ (lines 20-21). In this way, the counterfactual dataset \textit{D}$_{cf}$ is constructed while adhering to the constraint of minimal edits. However, additional operations are required to generate coreferential CAD from the original non-coreferential example. This is necessary to ensure that all event-relevant arguments co-refer with their counterparts within the pairwise context, as the definition of event coreference~\cite{cybulska2015bag}. Therefore, a new event-mention with discourse context, which co-refers to the second event-mention in \textit{MP} needs to be constructed. We begin with utilizing a paraphraser to generate prefix and suffix sentences based on those of the second event-mention (line 23-24) and then combine them with the mention sentence $s_{g}$ to construct the required event-mention text $\tilde{m}_{1}$ in line 25. In line 26, we pair $\tilde{m}_{1}$ with the original text of the second event-mention in the original \textit{MP}. Thus, the desired commonsense reasonable CAD is constructed with relatively minor text changes. 

The plausible counterfactual should ensure minimal changes compared with the original data. Otherwise, it may hurt the model's performance and robustness~\cite{keane2020good}. Inspired by~\citet{yang2022exploring}, we use MoverScore, an edit-distance scoring metric~\cite{zhao-etal-2019-moverscore}, to evaluate the plausibility of our generated counterfactual data. The average MoverScore of all CAD for ECB+ is 0.7339, which is much greater than the common plausibility baseline of 0.5. This demonstrates the minor changes in our CAD and validates the plausible quality of these generated instances.

\begin{algorithm}[tb]
    \small
    \caption{LLM-in-the-loop Counterfactual Generation}
    \label{alg:algorithm}
    \textbf{Input}: Original data \textit{MP} with label \textit{Y}; Large language model $LLM$; trigger terms of two mentions ($T^{\left( 1\right)}$,$T^{\left( 2\right)}$).\\
    \textbf{Prompt operators}: Synonyms generator $SYN$; Coref events generator $CE$; Non-coref events generator $NCE$; Paraphraser $PARA$.\\
    \textbf{Output}: Counterfactual dataset \textit{D}$_{cf}$
    \begin{algorithmic}[1] %[1] enables line numbers
    \WHILE{sentence $s$ in $MP$}
        \IF{$Y==coref$}
            \IF {$s==S^{\left( 1\right)  }_{i}$}
                \STATE $T^{\left( 1\right)  }_{syns}=LLM\left( SYN,T^{\left( 1\right)  }\right)$
                \STATE $S_{gens}=LLM\left( NCE,T^{\left( 1\right)  }_{syns}, S^{\left( 1\right)  }_{i}\right)$  
            \ELSE 
                \STATE continue
            \ENDIF
        \ELSIF{$Y==not\ coref$}
            \IF {$s==S^{\left( 2\right)  }_{i}$}
                \STATE $T^{\left( 2\right)  }_{syns}=LLM\left( SYN,T^{\left( 2\right)  }\right)$
                \STATE $S_{gens}=LLM\left( CE,T^{\left( 2\right)  }_{syns},S^{\left( 2\right)  }_{j}\right)$ 
            \ELSE 
                \STATE continue
            \ENDIF
        \ENDIF
    \ENDWHILE
    \WHILE{sentence $s_{g}$ in $S_{gens}$}
        \IF{$Y==coref$}
            \STATE $\tilde{m}_{1} =( S^{\left( 1\right)  }_{i-w},...,s_{g},...,S^{\left( 1\right)  }_{i+w})$  
            \STATE \textit{MP}$_{cf}=concat\left\{\tilde{m}_{1},\left( S^{\left( 2\right)  }_{j-w},...,S^{\left( 2\right)  }_{j+w}\right)  \right\}$  
        \ELSIF{$Y==not\ coref$}
            \STATE $pre=LLM\left( PARA,\  (S^{\left( 2\right)  }_{j-w},...,S^{\left( 2\right)  }_{j-1})\right)  $
            \STATE$suf=LLM\left( PARA,\  (S^{\left( 2\right)  }_{j+1},...,S^{\left( 2\right)  }_{j+w})\right)  $
            \STATE $\tilde{m}_{1} =concat\left\{  pre,s_{g},suf\right\} $ 
            \STATE \textit{MP}$_{cf}=concat\left\{\tilde{m}_{1},\left( S^{\left( 2\right)  }_{j-w},...,S^{\left( 2\right)  }_{j+w}\right)  \right\}$  
        \ENDIF
    \ENDWHILE
    \STATE Add \textit{MP}$_{cf}$ to the set \textit{D}$_{cf}$
    \STATE \textbf{return} \textit{D}$_{cf}$
    \end{algorithmic}
\end{algorithm}

\section{Experimental Settings}
\subsection{Evaluation Metrics and Datasets}
\label{sec:eval_metrics_datasets}
\textbf{Evaluation Metrics}
Since we do not conduct identification of the mention, we use B$^3$ F1 proposed by \citet{bagga-baldwin-1998-entity-based} to select the best model during training because \citet{moosavi-strube-2016-coreference} to identify that it has the fewest relevant drawbacks under the condition~\cite{held-etal-2021-focus}. For a comprehensive comparison with recent works, we also report MUC ~\cite{vilain-etal-1995-model}, CEAF$_{e}$ ~\cite{Ceafe}, LEA~\cite{moosavi-strube-2016-coreference} and CoNLL F1 which is the arithmetic average of the value of B$^3$, MUC and CEAF$_{e}$ F1.  

\noindent \textbf{Datasets} 
Our experiments are performed on three benchmarks: Event Coreference Bank Plus (ECB+), Football Coreference Corpus (FCC) and Gun Violence Corpus (GVC). For ECB+, we follow the data split by~\citet{cybulska2015bag}, while following the data split by~\citet{bugert-etal-2021-generalizing} for FCC and GVC. The data details are presented in Appx. Table~\ref{tab:datasets}.  
\subsection{Implementation Details}
\label{sec:implementation_details}
\textbf{LLM}
To evaluate the ECR performance of LLMs, we employ the document template prompt (Appx. Table~\ref{tab:doc_temp}) as suggested by~\citet{le2023large}. This prompt has shown considerably superior performance compared to the standard QA prompt and competes well with existing unsupervised entity coreference resolution systems. 
The evaluation is performed on the test set of ECB+. In practice, we begin by clustering the documents into golden topics, and subsequently, we evaluate the event coreference within each topic individually. ECB+ does not include cross-topic coreference links, so this operation will overlook incorrect coreference links across topics, thus simplifying the task. We do this to ensure that each prompt does not exceed the maximum acceptable length of GPT-4.

\noindent\textbf{Fully Fine-tuned Experiments}
To compare with the main baseline~\cite{held-etal-2021-focus} fairly, we follow their setup. For main experiments on three benchmarks, we retrieve the nearest 15 (\textit{K}=15) and 5 (\textit{K}=5) mention pairs for training and inference in main experiments on three benchmarks. For the ablation study and generalization test, we retrieve 5 (\textit{K}=5) mention pairs for both training and inference. Considering a trade-off between the training time and the increasing amount of augmented data, we only add two CAD for each original data from the top 5 nearest pairwise data in the training set, and keep the others unchanged. After data augmentation, we receive 68.2K, 35.8K and 97.3K mention pairs to train the cross-encoder on ECB+, FCC and GVC respectively. All of our models are trained and evaluated on a single Nvidia Tesla V100 GPU. All our augmented data originates from GPT-3.5-turbo~\cite{OpenAI}.

\section{Experimental Results}
\begin{table}[t]
    \small
    \centering
    \begin{tabular}{cc}
        \toprule
        \multirow{1}{*}{Methods} & \multicolumn{1}{c}{CoNLL F1} \\
        \midrule
        Lemma Matching~\cite{barhom2019revisiting}             & 76.5 \\
        E2E Neural System~\cite{cattan2020streamlining}        & 81.0 \\
        Pipeline System~\cite{held-etal-2021-focus}            & 85.7\\
        Causally Enhanced System (Ours)                               & \textbf{86.0} \\
        \midrule
        Claude-2              & 56.9$\downarrow$\\
        GPT-4                 & \underline{70$\downarrow$}\\
        \bottomrule
    \end{tabular}
    \caption{LLMs performance compared with other systems. The best overall result is highlighted in bold, while the best result among LLMs is underlined.}
    \label{tab:LLM_eval}
\end{table}
\textbf{LLMs}

Table~\ref{tab:LLM_eval} shows the cross-document ECR results by LLMs. Claude-2 lags significantly behind GPT-4 by 13.1 CoNLL F1 points. After checking the answers, we find that the low performance of Claude-2 is mainly attributed to the error type `Missing the golden mention', whose detail is presented in Appx. Sec~\ref{sec:Case_study_for_LLM}. Claude-2 misses 15\% of the total golden mentions of ECB+, including all golden mentions within Topic 37.
GPT-4 predicts more completely, with only 16 out of a total of 1780 golden event mentions being missed. Although GPT-4 performs better than Claude-2, it still falls short compared to other baselines. 
GPT-4's performance decreases by 6.5 CoNLL F1 points compared to the simple baseline~\cite{barhom2019revisiting}, which relies solely on event head lemma matching for coreference. Also, it falls significantly behind the current state-of-the-art pipeline method~\cite{held-etal-2021-focus}. In particular, our method outperforms GPT-4 with 16.0 CoNLL F1 points. The experimental results provide direct evidence that LLMs are not enough to solve the cross-document ECR problem and also demonstrate the effectiveness of our causally enhanced system based on LLM-RCDA. 
For further details related to the results of LLM evaluation, please refer to Appx. Sec~\ref{sec:Case_study_for_LLM}. 

\begin{table*}[t]
    \small
    \centering
    \resizebox{0.87\textwidth}{!}{
    \begin{tabular}{cccccccccccccc}
    \toprule
    \multirow{2}{*}{Methods} & \multicolumn{3}{c}{MUC}                             & \multicolumn{3}{c}{B$^3$}                                  & \multicolumn{3}{c}{CEAF$_{e}$}                                   & \multicolumn{3}{c}{LEA}                                     & CoNLL               \\ \cline{2-14} 
                             & R             & P             & F1                  & R                   & P                   & F1                   & R                   & P                   & F1                  & R                   & P                   & F1                  & F1                  \\ \midrule
    \textbf{ECB+}            &               &               &                     &                     &                     &                      &                     &                     &                     &                     &                     &                     &                     \\
    \citet{barhom2019revisiting}      & 77.6          & 84.5          & 80.9                & 76.1                & 85.1                & 80.3                 & 81.0                  & 73.8                & 77.3                & -                   & -                   & -                   & 79.5                \\
    \citet{cattan2020streamlining}            & 85.1          & 81.9          & 83.5                & 82.1                & 82.7                & 82.4                 & 75.2                & 78.9                & 77.0                  & -                   & -                   & -                   & 81.0                  \\
    \citet{bugert-etal-2021-generalizing}            & 76.0            & 76.1          & 76.1                & 71.8                & 81.2                & 76.2                 & 72.2                & 72.1                & 72.2                & 55.1                & 67.9                & 60.8                & 74.8                \\
    \citet{caciularu2021cdlm}                & 87.1          & \textbf{89.2} & \textbf{88.1}                & 84.9                & 87.9                & 86.4                 & 83.3                & 81.2                & 82.2                & 76.7                & 77.2                & 76.9                & 85.6                \\
    \citet{held-etal-2021-focus}                     & 87.0            & 88.1          & 87.5                & 85.6                & 87.7                & 86.6                 & 80.3                & \textbf{85.8}                & 82.9                & 74.9                & 73.2                & 74.0                  & 85.7                \\
    \citet{yu2023pairwise}                & \textbf{88.1} & 85.1          & 86.6                & \textbf{86.1}       & 84.7                & 85.4                 & 79.6                & 83.1       & 81.3                & -                   & -                   & -                   & 84.4                \\
    \citet{ahmed20232}                    & 80.0            & 87.3          & 83.5                & 79.6                & 85.4                & 82.4                 & 83.1                & 75.5                & 79.1                & 70.5                & 73.3                & 71.9                & 81.7                \\
    Baseline System$_{-DA}$             & 82.5          & 88.6          & 85.4                & 82.6                & \textbf{88.6}       & 85.5                 & \textbf{85.1}       & 78.5                & 81.7                & 74.0                  & 77.4                & 75.6                & 84.2                \\
    Enhanced System$_{+DA}$               & \underline{86.4}    & 88.6          & \underline{87.5} & \underline{85.7}          & 88.4                & \underline{\textbf{87.0}}   & 84.7                & \underline{82.2}          & \underline{\textbf{83.4}} & \underline{\textbf{77.4}} & \underline{\textbf{79.6}} & \underline{\textbf{78.5}} & \underline{\textbf{86.0*}}   \\ \midrule
    \textbf{FCC}             &               &               &                     &                     &                     &                      &                     &                     &                     &                     &                     &                     &                     \\
    \citet{barhom2019revisiting}      & -             & -             & -                   & 36.0                  & 83.0                  & 50.2                 & -                   & -                   & -                   & -                   & -                   & -                   & -                   \\
    \citet{bugert-etal-2021-generalizing}            & 82.7          & 78.3          & 80.4                & \textbf{70.8}       & 38.3                & 49.2                 & 28.2                & 40.4                & 33.2                & \textbf{60.4}       & 30.4                & 39.8                & 54.3                \\
    \citet{held-etal-2021-focus}                     & \textbf{86.4} & 75.7          & 80.7                & 61.6                & 65.4                & 63.5                 & 39.1                & 65.3                & 48.9                & 47.2                & 57                  & 51.6                & 64.4                \\
    Baseline System$_{-DA}$             & 79.2          & \textbf{88.9} & \textbf{83.7}       & 64.4                & 61.6                & 63.0                   & \textbf{73.3}       & 46.0                  & 56.5                & 58.1                & 47.2                & 52.1                & 67.7                \\
    Enhanced System$_{+DA}$               & 79.2          & 88.2          & 83.4                & \underline{66.8}          & \underline{\textbf{74.7}} & \underline{\textbf{70.5}} & 72.7                & \underline{\textbf{46.7}} & \underline{\textbf{56.9}} & \underline{60.1}          & \underline{\textbf{60.1}} & \underline{\textbf{60.1}} & \underline{\textbf{70.3}} \\ \midrule
    \textbf{GVC}             &               &               &                     &                     &                     &                      &                     &                     &                     &                     &                     &                     &                     \\
    \citet{barhom2019revisiting}      & -             & -             & -                   & 81.0                  & 66.0                  & 72.7                 & -                   & -                   & -                   & -                   & -                   & -                   & -                   \\
    \citet{bugert-etal-2021-generalizing}            & 66.3          & 78.1          & 71.7                & 49.9                & 73.6                & 59.5                 & 60.9                & 38.2                & 47.0                  & 38.2                & 56.5                & 45.6                & 59.4                \\
    \citet{held-etal-2021-focus}                     & \textbf{91.8} & 91.2          & \textbf{91.5}       & 82.2                & 83.8                & 83.0                   & 75.5                & \textbf{77.9}       & \textbf{76.7}       & 79.0                  & \textbf{82.3}       & \textbf{80.6}       & 83.7                \\
    \citet{ahmed20232}                    & 84.0            & 91.1          & 87.4                & 79.0                  & 76.4                & 77.7                 & 69.6                & 52.5                & 59.9                & 74.1                & 63.9                & 68.6                & 75.0                  \\
    Baseline System$_{-DA}$             & 89.3          & \textbf{92.3} & 90.8                & 82.1                & 85.7                & 83.8                 & 76.6                & 67.5                & 71.7                & 76.9                & 78.8                & 77.8                & 82.1                \\
    Enhanced System$_{+DA}$               & \underline{90.4}    & 92.1          & \underline{91.3}          & \underline{\textbf{84.8}} & \underline{\textbf{86.8}} & \underline{\textbf{85.8}} & \underline{\textbf{78.9}} & \underline{73.2}          & \underline{76.0}            & \underline{\textbf{79.8}} & \underline{80.7}          & \underline{80.2}          & \underline{\textbf{84.4}} \\ \bottomrule
    \end{tabular}
    }
    \caption{Performance comparison of different cross-document ECR systems on ECB+, FCC and GVC. Baseline System$_{-DA}$ results are obtained by reproducing the work of~\citeauthor{held-etal-2021-focus} without data augmentation. Enhanced System$_{+DA}$ is trained by the original data combined with CAD from LLM-RCDA. Bold values represent the overall best results, while underlined values indicate results that beat the Baseline System. {*} indicates the result is statistically different from the baseline result with $p<0.01$ via the pairwise-t test.}
    \label{tab:main_table}
\end{table*}

\noindent \textbf{Causally Enhanced System} 
As shown in Table~\ref{tab:main_table}, our causally enhanced ECR system has achieved state-of-the-art performance across multiple evaluation metrics. In terms of CoNLL F1, the system surpasses the baseline by 1.8, 2.6, and 2.3 points on ECB+, FCC, and GVC, respectively. 

In the case of ECB+, we observe a significant improvement in Recall for our enhanced system compared to the baseline system, as measured by MUC, B$^{3}$, and LEA, with an average improvement of 3.5 points. This improvement can be attributed to the trigger intervention in Algo.~\ref{alg:algorithm}. The introduction of diverse trigger expressions enhances the model's comprehension of event semantics, thereby rectifying false negatives caused solely by literal differences in trigger terms. The case study is presented in Appx. Sec~\ref{sec:Case_study_for_model_interpretability}. 

FCC and GVC represent single-domain datasets focused on football and gun violence news. They include a substantial volume of coreferential event mentions across various topics, resulting in a considerable number of challenging negatives (i.e., non-coreferential event-mention pairs with very similar contexts). Nevertheless, several metrics exhibit notable enhancements in Precision, such as a 13.1-point increase for B$^{3}$ on FCC and a 5.7-point increase for CEAF$_{e}$ on GVC. These results suggest that our LLM-RCDA method is well-suited for such scenarios, as it guides the model to make decisions based on fine-grained causal terms within the context.

\section{Analysis}
\label{sec:analysis}
\begin{table*}[t]
    % \small
    \centering
    \resizebox{1.0\textwidth}{!}{
    \begin{tabular}{cccccccccccccccc}
    \toprule
    \multirow{2}{*}{\begin{tabular}[c]{@{}c@{}}Training Data \\ (data volume)\end{tabular}} & \multirow{2}{*}{Percentage} & \multirow{2}{*}{MoverScore} & \multicolumn{3}{c}{MUC}                       & \multicolumn{3}{c}{B$^{3}$}                  & \multicolumn{3}{c}{CEAF$_{e}$}                   & \multicolumn{3}{c}{LEA}                       & CoNLL       \\ \cline{4-16} 
                                                                                            &                             &                             & R             & P             & F1            & R             & P             & F1            & R             & P             & F1            & R             & P             & F1            & F1          \\ \midrule
    ORI$_{-TI/-CI}$ (14.3K)                                                                             & 70.0\%                      & -                           & 83.0            & 85.9          & 84.4          & 84.2          & 85.7          & 85.0            & 81.9          & 78.7          & 80.2          & 74.6          & 74.2          & 74.4          & 83.2        \\
    
    ORI\&TAD$_{+ TCDA}$(42.8K)                                                                            & 19.9\%                      & 0.5457
          & 75.7          & \textbf{89.0}            & 81.8          & 78.2            & \textbf{89.8}          & 83.6        & \textbf{86.3}            &73.0           & 79.1          &
    69.7          & 75.2          & 72.4          & 81.5$\downarrow$          \\
    
    ORI\&TIA$_{-TI/+CI}$ (42.8K)                                                                        & 91.4\%                      & 0.7354                      & 82.1          & 85.7          & 83.9          & 83.3          & 85.2          & 84.2          & 82.7 & 78.6          & 80.6          & 73.9          & 73.7          & 73.8          & 82.9$\downarrow$        \\
    ORI\&CIA$_{+TI/-CI}$ (42.8K)                                                                        & 19.9\%                      & 0.6928                      & 84.1          & 86.2          & 85.2          & 84.5          & 86.5 & 85.5          & 82.4          & 80.0            & 81.2          & 74.7          & 75.5          & 75.1          & 84.0$\uparrow$          \\
    ORI\&CAD$_{+TI/+CI}$ (42.8K)                                                                        & 19.9\%                      & 0.7339                      & \textbf{87.2} & 86.5 & \textbf{86.8} & \textbf{86.7} & 85.4          & \textbf{86.1} & 81.8          & \textbf{82.7} & \textbf{82.2} & \textbf{77.2} & \textbf{76.2} & \textbf{76.7} & \textbf{85.0$\uparrow$} \\ \bottomrule
    \end{tabular}
    }
    \caption{Results of the system trained with different data combinations on ECB+. Percentage denotes the proportion of examples with lexically similar triggers in coreferential pairs. $\cdot_{+/-TI(CI)}$ indicates whether the Trigger (Context) Intervention is included or excluded. $\cdot_{+TCDA}$ means that the augmented data is from~\citeauthor{ravi2023happens}'s TCDA method.}
    \label{tab:ablation_study}
\end{table*}

\textbf{Ablation Study}
Our LLM-RCDA algorithm assists the cross-document ECR system in disentangling spurious patterns via Trigger Intervention (TI) while emphasizing causal associations through Context Intervention (CI) when understanding the pairwise context. To investigate the efficacy of each intervention, we modify the augmented data generation algorithm (Algo. ~\ref{alg:algorithm}) to conduct the ablation study for TI and CI. For TI ablation (Appx. Algo.~\ref{alg:tia} ), we no longer diversify the expressions of the target mention's trigger, resulting in trigger pairs in the augmented data that remain consistent with those in the original mention pair. For CI ablation (Appx. Algo.~\ref{alg:cia}), we intentionally introduced more substantial modifications to the text of the original mention pair. This deliberate approach leads to the generation of relatively implausible counterfactuals, aiming to reduce the emphasis on rationales within the context~\cite{keane2020good,yang2022exploring}. The augmented data from TI and CI ablation are short for TIA and CIA, while the augmented data from the complete LLM-RCDA pipeline is short for CAD. Table~\ref{tab:ablation_study} compares results from different data combinations. Overall, ORI\&CAD outperforms other data combinations across multiple metrics.

We first discuss the necessity of TI. The absence of TI in ORI\&TIA leads to approximately 92\% of coreferential training data involving lexically similar trigger pairs (22\% higher than that of ORI), which inevitably exacerbates the `triggers lexical matching' bias. Consequently, ORI\&TIA experiences a performance decline of 0.3 CoNLL F1 compared to ORI, despite increasing the training data volume by threefold through the addition of plausible counterfactuals. However, with the help of TI, ORI\&CAD shows greater lexical variation in the literal surface forms of trigger terms, with only 19.9\% (v.s., 91.4\% for ORI\&TIA and 70.0\% for ORI) of the coreferential data containing lexically similar trigger pairs. This data combination leads to a substantial performance improvement in the ECR system, surpassing ORI by 1.8 CoNLL F1 and ORI\&TIA by 2.1 CoNLL F1, which indicates that mitigating `triggers lexical matching' bias through TI can lead to performance enhancements. 

Furthermore, the performance comparison between ORI\&CIA and ORI\&CAD demonstrates the necessity of utilizing CI. Due to the relaxation of the minimum edit distance constraint, CIA becomes the implausible counterfactual data, with a MoverScore of 0.6928 which is lower than 0.7339 of CAD. The performance of ORI\&CIA falls behind that of ORI\&CAD by 1.0 CoNLL F1. This implies that CI ablation weakens the model's ability to capture and understand rationales, resulting in only sub-optimal performance.

The ablation study highlights the importance of both TI and CI and validates our analysis of spurious and causal associations. Therefore, the most effective way to enhance cross-document ECR performance is by fully utilizing our LLM-RCDA algorithm.  

\noindent \textbf{Comparison with Temporal Commonsense DA}
~\citeauthor{ravi2023happens} enriched the event context by introducing possible preceding or succeeding scenarios related to event mentions based on the Temporal Commonsense Event Coreference Data Augmentation (TCDA) method, thereby increasing the distinction between events. They also designed an inference-enhanced pairwise scorer specifically to capture such temporal information. We leverage their prompt to generate the temporal commonsense augmented data (TAD) and then incorporate them into the original dataset (ORI) to train our baseline ECR system~\cite{held-etal-2021-focus}  (Appx. Algo.~\ref{alg:tad}). Results in Table~\ref{tab:ablation_study} exhibit that ORI\&CAD outperforms ORI\&TAD across various metrics, improving by 3.5 CoNLL F1. Also, the performance of ORI\&TAD is worse than that of the system trained solely with ORI. This observation indicates that the TCDA method heavily relies on their tailored scorer. In contrast, LLM-RCDA improves performance without requiring system modifications, with more convenience and scalability.

\noindent \textbf{Robustness in the Generalization Test}
\begin{table}[t]
    \centering
    \resizebox{1.0\columnwidth}{!}{
    \begin{tabular}{cccccc}
        \toprule
        \multirow{1}{*}{Methods} & \multicolumn{1}{c}{MUC} & \multicolumn{1}{c}{B$^3$}  & \multicolumn{1}{c}{CEAFe} & {LEA} & {CoNLL}\\
        \midrule
        \citet{bugert-etal-2021-generalizing}         & 52.4 & 33.2 & 27.0 & 14.1 & 33.2 \\
        Baseline System~\cite{held-etal-2021-focus}                   & 57.5 & 38.4 & 31.5 & 23.9 & 42.5 \\
        Enhanced System (ours) & \textbf{68.6} & \textbf{46.0} & \textbf{35.0} &\textbf{ 31.2} & \textbf{49.9}\\
        \bottomrule
    \end{tabular}
    }
    \caption{Performance comparison of our enhanced system with baselines on the OOD dataset FCC.}
    \label{tab:cross_corpus}
\end{table}
We train the system on ECB+ but test it on FCC to evaluate its out-of-the-domain (OOD) robustness. For comparison, we take the cross-corpus results from~\citet{bugert-etal-2021-generalizing} and the reproduced results from~\citet{held-etal-2021-focus} as our baselines. The enhanced system is trained with ORI and CAD from LLM-RCDA. As shown in Table~\ref{tab:cross_corpus}, our enhanced system shows the best performance in multiple metrics. It surpasses the baseline system by 7.4 CoNLL F1 points, proving the stronger robustness of LLM-RCDA.

To better demonstrate how our enhanced system performs more robustly in the OOD scenario, we perform an error analysis. We randomly sample 100 errors made by the baseline system but correctly predicted by our enhanced ECR system, including 50 false positives (FPs) and 50 false negatives (FNs). FNs refer to coreferential examples being incorrectly predicted as non-coreferential, while FPs refer to the non-coreferential examples being wrongly predicted as coreferential. According to the context of these mention pairs, we manually categorize them into five error types: `Ignore argument counterparts', `Require contextual understanding', `Lack of the evidence', `Without domain expertise' and `Annotation mistakes' and analyze the distribution of them.

\begin{figure}[t]
    \centering
    \includegraphics[width=1.0\columnwidth]{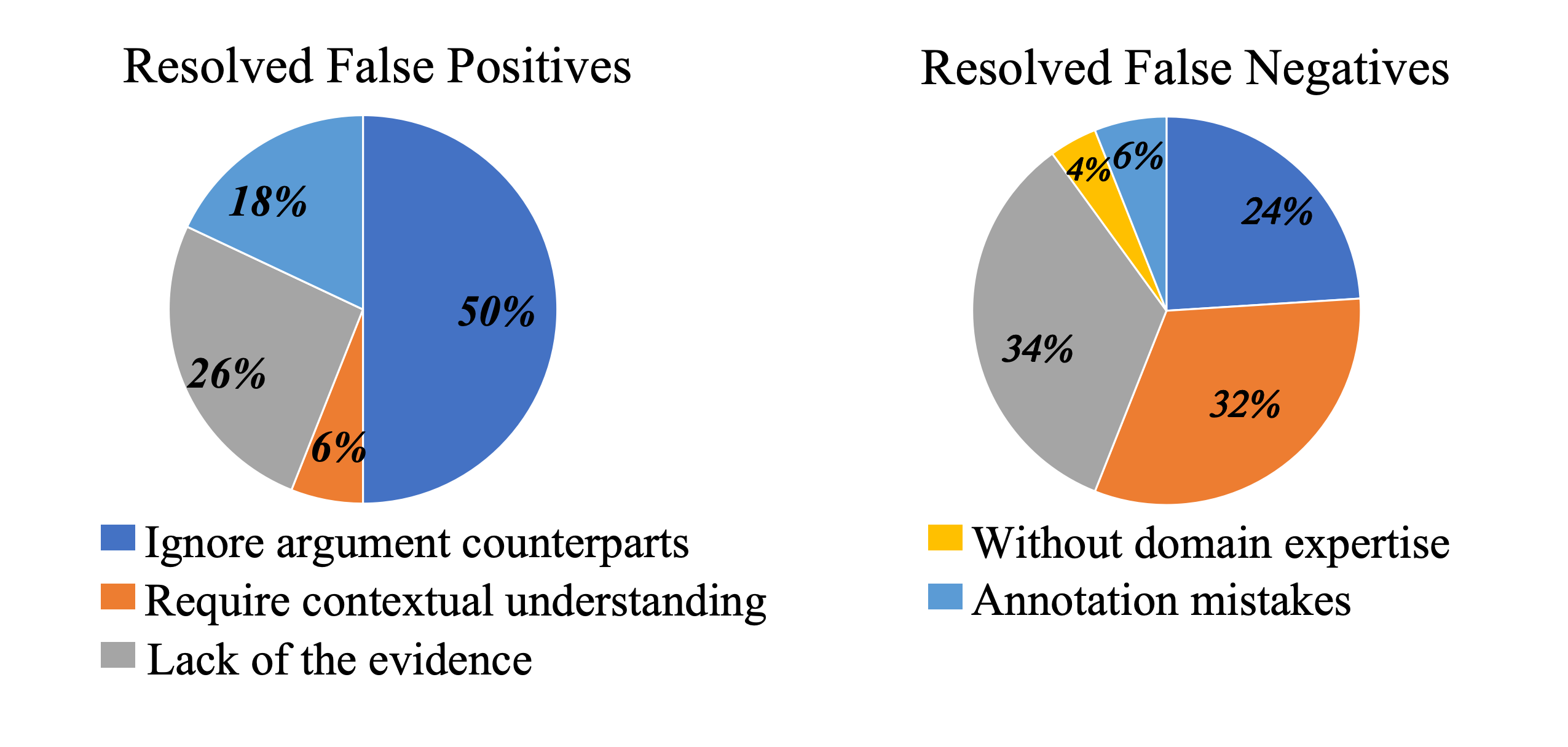} 
    \caption{Error distribution on resolved baseline errors by the enhanced system.}
    \label{fig:error_analysis}
\end{figure}

From Figure~\ref{fig:error_analysis}, we observe that 50\% of FPs fall under the category of `Ignore argument counterparts', which represents simple cases for the ECR system. In these samples, the coreference of event-relevant arguments (i.e., rationale) is clearly present within the sentences where the event mention occurs. For example: 
\begin{center}
    \small
    \fbox{\parbox{0.9\linewidth}{
    1. ...\textbf{\textit{Defeated}} Denmark in the round of 16 in a penalty shootout, 3-2, after drawing 1-1...\\
    2. ...\textbf{\textit{Defeated}} Spain in the round of 16 in a penalty shootout, 4-3, after drawing 1-1...}
    }
\end{center}
This type of error would be avoided if the system captured non-coreferential evidence from the participant argument counterpart `\textit{Denmark}' and `\textit{Spain}'. The enhanced system's ability to correctly handle such examples may be attributed to the counterfactual data (CAD) for original positive instances (shown in Appx. Table~\ref{tab:cad_tia_cia_example}). Training with such data enhances the model's sensitivity to the surrounding arguments of the trigger.

Among FNs, `Ignore argument counterparts' remains an important component, accounting for 24\%. Unlike the same error type in FPs, the baseline system fails to capture coreferential evidence from argument counterparts in the pairwise context. Besides, `Require contextual understanding' takes 32\% of resolved FNs. Such errors pose greater challenges, as their pairwise context exhibits significant syntactic structural differences. While contexts provide ample information about event-relevant arguments, accurately identifying and understanding the relationships between these arguments is not easy for the baseline system. For example:
\begin{center}
    \small
    \fbox{\parbox{0.9\linewidth}{
    1. ...Belgium defeated England 2-0 in the World Cup's third-place game in St. Petersburg. The victory meant Belgium, who \textbf{\textit{lost}} to France in the semifinals on Tuesday...\\
    2. ...Belgium and England will contest the third-place play-off at the St. Petersburg Stadium on Saturday. The Red Devils \textbf{\textit{lost}} 1-0 to France before England succumbed to a Croatia comeback...}
    }
\end{center}
To give a correct answer, the ECR system needs to infer, based on the context, that the antecedent `\textit{who}' in the first text snippet refers to `\textit{Belgium}', and `\textit{The Red Devils}' in the second text snippet refers to `\textit{Belgium}' as well. Then, it needs to link the coreferential relationship between counterpart arguments to conclude that these two mentions co-refer to the event `\textit{Belgium lost to France}'. Examples and analyses of other error types can be found in Appx. Sec~\ref{sec:Case_study_for_generalization_test}. 

The error analysis reveals a deficiency in the baseline system's comprehension of complete semantics. In contrast, our enhanced ECR system effectively addresses these errors. This is attributed to LLM-RCDA, which enables a rationale-centric decision in solving ECR by enhancing the system's ability to identify and understand the evidence of event-relevant arguments within the pairwise context, which is the key to prediction.

\section{Conclusion}
We proposed a novel rationale-centric counterfactual data augmentation method specialized for the pairwise text input of cross-document ECR system, which leverages interventions from LLM to enhance the system's event coreference decision causally. Experimental results verify the significant performance and robustness improvement of the enhanced ECR system with our method. 

\section*{Limitations}
The LLM used in our LLM-RCDA method is GPT-3.5-turbo~\cite{OpenAI}, and it is not an open-source model. In the future, we plan to attempt to implement our method based on some open-source large models, such as LLaMA~\cite{touvron2023llama}. Additionally, we aim to apply it to other cross-document tasks, not limited to event coreference resolution. We are also interested in adapting our method to other pairwise input text tasks, such as natural language inference, stance detection, and entity coreference resolution.

\section*{Ethical Statement}
We honour the Code of Ethics of ACL. 

\section*{Acknowledgements}
We thank the anonymous reviewers for their insightful comments and suggestions to help improve the paper. This publication has emanated from research conducted with the financial support of the Pioneer
and `Leading Goose' R\&D Program of Zhejiang under Grant Number 2022SDXHDX0003 and the National Natural Science Foundation of China Key Program under Grant Number 62336006.

\bibliography{acl_latex}
\bibliographystyle{acl_natbib}
\clearpage

\appendix

\section{Appendix}
\label{sec:appendix}
\begin{table}[H]
    \small
    \centering
    \begin{tabular}{cccc}
    \toprule
    \textbf{ECB+}  & Train   & Dev    & Test    \\ \midrule
    Topics         & 25 (50) & 8 (16) & 10 (20) \\
    Documents      & 574     & 196    & 206     \\
    Sentences      & 9366    & 2837   & 3505    \\
    Event Mentions & 3808    & 1245   & 1780    \\ \midrule
    \textbf{FCC}   & Train   & Dev    & Test    \\ \midrule
    Topics         & 3       & 1      & 1       \\
    Documents      & 207     & 117    & 127     \\
    Sentences      & 7018    & 3648   & 4274    \\
    Event Mentions & 1604    & 680    & 1074    \\ \midrule
    \textbf{GVC}   & Train   & Dev    & Test    \\ \midrule
    Topics         & 1 (170) & 1 (37) & 1 (34)  \\
    Documents      & 358     & 78     & 74      \\
    Sentences      & 7607    & 1325   & 1360    \\
    Event Mentions & 5313    & 977    & 1008    \\ \bottomrule
    \end{tabular}
    \caption{Statistics for ECB+, FCC and GVC. For Topics rows, values outside the parentheses indicate the number of topics, while values inside the parentheses represent the number of subtopics of the data split (e.g., 25 (50) means that 25 topics including 50 subtopics are in the data split).}
    \label{tab:datasets}
\end{table}

\subsection{Dataset Details}
\label{sec:dataset_details}
\textbf{Event Coreference Bank Plus (ECB+)} 
The ECB+ corpus is the most popular benchmark for the cross-document ECR task~\cite{cybulska-vossen-2014-using}. It is an extension of the Event Coref Bank corpus (ECB) annotated by~\citet{bejan-harabagiu-2010-unsupervised}. ECB+ expands on the original topics by incorporating various seminal events as subtopics and annotates the coreference relationships between events within each topic. In terms of statistics, the ECB+ corpus consists of 982 documents, covering 43 topics, and includes 26,712 coreference links among 6,833 event mentions.
\\
\textbf{Football Coreference Corpus (FCC)}
The Football Coreference Corpus (FCC) serves as a benchmark for cross-document Event Coreference Resolution (ECR) specifically in the domain of football tournaments~\cite{bugert-etal-2021-generalizing}. This dataset is unique as it includes a significant number of cross-subtopic event coreference links, which is uncommon but highly valuable for research purposes. Overall, the FCC comprises 451 documents and contains a total of 145,272 links between 3,563 event mentions.
\\
\textbf{Gun Violence Corpus (GVC)}
The GVC~\cite{vossen-etal-2018-dont} is a challenging cross-document ECR benchmark. It consists of 510 documents that are lexically similar, posing a challenge for document clustering. The dataset comprises 29,398 links between 7,298 event mentions, with all the links being within subtopics.

\subsection{Experimental Details}
\label{sec:experimental_details}
\subsubsection{Triggers lexical similarity}
\label{sec:Triggers_lexical_similarity}
The fuzz ratio leverages Levenshtein Distance to calculate the differences between sequences, enabling us to intuitively measure the lexical similarity between terms. A ratio closer to 100 indicates greater lexical matching between two trigger terms, while a lower value suggests lexical divergence. 

In our work, we use package \textit{thefuzz}~\footnotemark[1] to calculate the fuzz ratio between triggers' etymas for the measurement. If the ratio is greater than or equal to 80, the example is considered lexically similar; otherwise, it is deemed lexically divergent. Therefore, we calculate the average triggers lexical similarity of coreferential examples for different data combinations in Table~\ref{tab:ablation_study}: 78.3 for ORI; 93.84 for ORI\&TIA; 40.6 for ORI\&TAD, ORI\&CIA and ORI\&CAD. This indicates that `Trigger intervention' can make trigger pairs more diverse in terms of their lexical forms. 

\subsubsection{LLM Evaluation}
\label{sec:llm_evaluation}
\footnotetext[1]{https://github.com/seatgeek/thefuzz}
\footnotetext[2]{https://platform.openai.com}
\footnotetext[3]{https://www.anthropic.com}
We utilize MUC, B$^{3}$, CEAF$_{e}$, LEA and CoNLL, five metrics to evaluate the performance of Claude-2 and GPT-4 on cross-document ECR. Claude-2 accepts 100K input, but we have no access to adjust the parameters currently, so we interact with it directly on its official website~\footnotemark[3]. The GPT-4 we used accepts 8K tokens shared between the prompt and output in maximum~\footnotemark[2]. We set the temperature parameter as zero for reproducibility, and adjust the maximum length as 4500. Other parameters are set as default.
\subsubsection{Representation}
\label{sec:representation}
Following~\citet{held-etal-2021-focus}, we arrange a text snippet for each mention by including sentences from a context window preceding and following the mention sentence. Subsequently, a fine-tuned bi-encoder initialized with pre-trained weights from RoBERTA-large (326M)~\cite{liu2019roberta} encodes the token-level boundary representation used by~\cite{lee-etal-2017-end}. The mention is then represented by concatenating these token-level representations. As a result, the pairwise representation for the mention pair can be constructed as follows:
$$\left[ m_{1},m_{2},m_{1}\odot m_{2}\right]  $$
where $m_{1}$ and $m_{2}$ refer to the representation of the first and second mention in the pair, and $m_{1}\odot m_{2}$ denotes the element-wise multiplication between two mention. 

The original data, along with the augmented data, is encoded into the pairwise representation and then passed through a cross-encoder for training a coreference classifier.
\subsubsection{Hyperparameters}
\label{sec:hyperparam}
Hyperparameters for training the bi-encoder and cross-encoder are presented in Table~\ref{tab:hps}. 

\begin{table}[t]
    \small
    \centering
    \begin{tabular}{ccc}
    \toprule
                        & bi-encoder & cross-encoder \\ \midrule
    Batch Size          & 16         & 40 (8)        \\
    Learning Rate       & 1e-5    &  1e-5       \\
    Maximum Epochs      & 50         & 40            \\
    Optimizer           & Adam       & Adam          \\
    Warmup Proportion   & 0.1        & 0.1           \\
    Early Stop Patience & 10         & 5             \\
    Max Grad Norm       & 1          & 1             \\
    Input Turncation    & 512        & 512           \\ \bottomrule
    \end{tabular}
    \caption{Hyperparameters settings. We set the batch size as 40, 40, 8 for cross-encoder when performing corpus-tailored study on ECB+, FCC and GVC respectively, while setting it as 8 for ablation and generalization study.}
    \label{tab:hps}
\end{table}

\subsection{Prompts}
\label{sec:prompts}
\subsubsection{The prompt for LLM evaluation}
\label{sec:prompt_for_llm_eval}
To evaluate the performance of LLM on the cross-document ECR, we follow and modify the document template prompt (Table~\ref{tab:doc_temp}) proposed by~\citet{le2023large}, which outperforms several existing unsupervised coreference systems in the entity coreference resolution task on the OntoNotes dataset~\cite{le2023large}.
\begin{table*}[t]
    \resizebox{\linewidth}{!}{
        \begin{tabular}{l}
            \toprule
            \begin{tabular}[c]{@{}l@{}}
    \textbf{Document Template prompt:}\\Annotate all event mentions in the following text with coreference clusters. Use Markdown tags to indicate clusters in the output, with the \\following format [mention](\#cluster\_name):\\(...) Lindsay Lohan [checks into](\#) Betty Ford Center May 03, 2013. After skipping out on [entering](\#) a Newport Beach rehabilitation \\facility and [facing](\#) the prospect of [arrest](\#) for [violating](\#) her [probation](\#), Lindsay Lohan has [checked into](\#) the Betty Ford \\Center to [begin](\#) a 90-day court-mandated [stay](\#) in her reckless [driving](\#) [conviction](\#). (...)~\textbf{}\\\textbf{LLM Response:}\\(...) Lindsay Lohan [checks into](\#cluster1) Betty Ford Center May 03, 2013. After skipping out on [entering](\#cluster2) a Newport Beach \\rehabilitation facility and [facing](\#cluster3) the prospect of [arrest](\#cluster4) for [violating](\#cluster5) her [probation](\#cluster6), Lindsay \\Lohan has [checked into](\#cluster1) the Betty Ford Center to [begin](\#cluster7) a 90-day court-mandated [stay](\#cluster8) in her reckless\\{[}driving](\#cluster9) [conviction](\#cluster10). (...)\end{tabular}  \\
    \bottomrule
    \end{tabular}}
    \caption{An example of \textbf{Doc}ument \textbf{Temp}late (DocTemp) prompt for event coreference resolution, where mention head lemmas are marked with [mention](\#), allowing the LLM to provide the coreference cluster IDs for based on the semantics of the entire document~\cite{le2023large}.}
    \label{tab:doc_temp}
\end{table*}

\begin{table*}[t]
    \centering
    \resizebox{1.8\columnwidth}{!}{
        \begin{tabular}{cccccccccccccc}
        \toprule
        \multirow{2}{*}{Methods}                  & \multicolumn{3}{c}{MUC}                     & \multicolumn{3}{c}{B$^{3}$}                   & \multicolumn{3}{c}{CEAF$_{e}$}                & \multicolumn{3}{c}{LEA}                     & CoNLL         \\ \cline{2-14} 
                                                  & R           & P             & F1            & R             & P             & F1            & R             & P             & F1            & R           & P             & F1            & F1            \\ \midrule
        Claude-2                      & 39.0          & 61.8          & 47.8          & 51.6          & \textbf{76.3}          & 61.6          & \textbf{65.0}            & 58.0            & 61.3          & 65.0          & 78.3          & 71.0            & 56.9          \\
        GPT-4                          & \textbf{73.5}        & \textbf{73.5}          & \textbf{73.5}          & \textbf{79.8} & 67.8          & \textbf{73.3}          & 62.4          & \textbf{63.8}          & \textbf{63.1}          & \textbf{65.0}          & \textbf{86.4}          & \textbf{74.2}          & \textbf{70.0}            \\ \bottomrule
    \end{tabular}}
    \caption{A comprehensive LLMs evaluation based on DocTemp prompt. The overall best results are bolded.}
    \label{tab:LLM_eval_zero_shot}
\end{table*}

\subsubsection{The prompt for data generation} 
\label{sec:prompt_for_data_gen}
Algorithm~\ref{alg:algorithm} contains the following prompt-based operations: generating the synonyms $SYN$ and
non-coreferential(coreferential) event mention sentence $NCE$($CE$), paraphrasing the given text $PARA$. From Table~\ref{tab:coref_syn_cf_prompt}, prompt Step 1 illustrates the operation \textit{SYN} in line 4, while prompt Step 2 corresponds to the operation \textit{NCE} in line 12. Meanwhile, the prompt Step 1 and Step 2 in Table~\ref{tab:non_coref_syn_cf_prompt} refer to the operation $SYN$ and $CE$ in lines 11 and 12 respectively. Table~\ref{tab:para} shows the $PARA$ prompt about paraphrasing the discourse context, corresponding to lines 22 and 23. Table~\ref{tab:tc_prompt} displays the prompts used to generate TAD, which is utilized in Algorithm~\ref{alg:tad} in the `Comparison with Temporal Commonsense DA' subsection of Section~\ref{sec:analysis}.
\subsection{Data Examples}
\label{sec:data_examples}
Table~\ref{tab:cad_tia_cia_example} and~\ref{tab:non_coref_cad_tia_cia_example} show examples of Counterfactual Augmented Data (CAD), Trigger Intervention Ablation (TIA) Data, Context Intervention Ablation (CIA) Data as well as the Temporal commonsense Argumented Data (TAD), where the special token `<s>' and `<$\backslash{}$s>' indicate the start and end of the sentence. 
The average MoverScore for CAD and TIA are both approximately 0.73, whereas for CIA it is around 0.69, which indicates CAD and TIA are more plausible counterfactual data.~\citet{yang2022factmix} have demonstrated that the performance and robustness of the model cannot be infinitely improved by adding more counterfactuals. Therefore, in our data augmentation process, we randomly select two CAD from counterfactual candidates generated by Algorithm 1 for each specified original data. To ensure fair comparison, we also two TIA and CIA instances for each original data in the ablation study. The generation process of TIA, CIA and TAD are presented in algorithm~\ref{alg:tia}, ~\ref{alg:cia} and~\ref{alg:tad} respectively.

\subsection{Additive experiments}
\subsubsection{Comparison with other counterfactual data generation method}
\label{sec:Comparison_other_CAD}
We implement an automatic counterfactual data augmentation technique named RM-CT (\textbf{r}e\textbf{m}oving the \textbf{c}ausal \textbf{t}erm) for comparison, which is originally designed for the sentiment analysis task~\cite{yang2022exploring}. Our implementation of the ECR RM-CT takes two steps:
(1) \textit{Identify the causal term (i.e., rationales).} We use the same semantic role labelling (SRL) tool used by~\citeauthor{yu2023pairwise}, to identify and parse all event-relevant arguments for a given event. These arguments (including event triggers) are rationales for the ECR task. (2) \textit{CAD generation.} For each original coreferential mention-pair, we randomly replace rationales in the first mention with the `<unk>' token to remove the rationale and destroy the coreference of the counterpart arguments, while for each original non-coreferential mention-pair, we randomly retrieve the mention coreferential to the second mention of the mention-pair from the training corpus, to construct its CAD.

The experimental setting is the same as that of the ablation study, which is mentioned in section~\ref{sec:implementation_details}. We still assign two augmented data for each raw data. 
\begin{table}[H]
    \centering
        \resizebox{1.0\columnwidth}{!}{
            \begin{tabular}{cccccc}
            \toprule
            \begin{tabular}[c]{@{}c@{}}Training data\\ (data volume)\end{tabular} & \multicolumn{1}{l}{MUC} & B$^{3}$      & CEAF$_{e}$    & LEA           & \multicolumn{1}{l}{CoNLL} \\ \midrule
            w/o DA (14.3K)                                                        & 84.4                    & 85.0            & 80.2          & 74.4          & 83.2                      \\
            RM-CT (42.8K)                                                         & 84.3                    & 82.3          & 76.6          & 70.3          & 81.1                      \\
            \multicolumn{1}{l}{LLM-RCDA (42.8K)}                                  & \textbf{86.8}           & \textbf{86.1} & \textbf{82.2} & \textbf{76.7} & \textbf{85.0}               \\ \bottomrule
            \end{tabular}
        }
    \caption{Performance comparison of two different CAD generation techniques.}
    \label{tab:RM-CT_vs_LLM-RCDA}
\end{table}
The results demonstrate that LLM-RCDA outperforms RM-CT across multiple metrics. Surprisingly, the introduction of RM-CT augmented data even results in a side effect, causing a deterioration in the performance of the original model trained without data augmentation. The robust performance advantage of LLM-RCDA can be attributed to several key strengths: (1) LLM-RCDA is more user-friendly. It does not require prior parsing of rationales using the SRL tool, making data preprocessing less cumbersome. Additionally, many popular SRL tools (e.g., AllenNLP SRL~\cite{gardner-etal-2018-allennlp}) struggle to parse event-relevant arguments for non-verb triggers, making it challenging to generate satisfied CAD for mention pairs containing non-verb triggers. (2) LLM-RCDA ensures more meaningful CAD for ECR. Leveraging the powerful controllable generation capability of LLM~\cite{li2023large,chen-etal-2023-disco}, we can effortlessly generate mentions which are coreferential or non-coreferential to a specified. This allows for easy flipping of the label of the original mention-pair with minimal changes in the text. In contrast, RM-CT attempts to directly flip the original label by simply removing an event-relevant argument. This is not suitable for ECR because the text describing an event mention may not always be accompanied by all event-relevant arguments. Even though some of the rationales of a mention-pair are removed, the label of the mention pair should still be unchanged by commonsense for most cases.

Consequently, employing RM-CT to generate CAD for ECR tasks seems arbitrary and ineffective for ECR and LLM-RCDA is a better solution.

\subsubsection{The influence of augmentation ratios}
\label{sec:influence_of_augmentation_ratios}
\begin{table}[H]
    \centering
        \resizebox{1.0\columnwidth}{!}{
            \begin{tabular}{cccc}
            \toprule
            \begin{tabular}[c]{@{}c@{}}Training data \\ (data volume)\end{tabular} & B$^{3}$ Recall & B$^{3}$ Precision & B$^{3}$ F1 \\ \midrule
            w/o DA (6.8K)                                                          & \textbf{64.1}            & 55.8               & 59.6        \\
            Ratio 1 DA (13.6K)                                                     & 61.1            & 68.1               & \textbf{64.4}        \\
            Ratio 3 DA (27.3K)                                                     & 60.0            & 66.8               & 63.2        \\
            Ratio 5 DA (40.8K)                                                     & 55.9            & \textbf{71.1}               & 62.8        \\ \bottomrule
            \end{tabular}
        }
    \caption{The influence of different augmentation ratios.}
    \label{tab: AD_ratios}
\end{table}
We implement an experiment to explore the existence of the plateau effect when adding more CAD. We construct the mention-pair dataset on FCC by retrieving the nearest five event mentions to pair with each raw mention in the training/development/test corpus. Subsequently, we vary the augmented ratio as 1, 3, and 5, signifying the addition of 1, 3, and 5 augmented data respectively, to the training set for each corresponding original data. The experimental setup here is the same as the corpus-tailored study conducted on the FCC dataset and B$^{3}$ F1 is used for selecting the best model. Experimental results are shown in Table~\ref{tab: AD_ratios}.

In terms of B$^{3}$ F1, a huge performance improvement over the baseline without DA when the augmented ratio is set as 1, which shows a 4.8 points improvement. When we add more CAD, such as when the augmented ratio is 3 or 5, performance improvement seems to reach a plateau, and the performance may even slightly decline. This plateau effect has also been observed and discussed in the FactMix~\cite{yang2022factmix}. One possible reason for this phenomenon is that excessively increasing the augmented data may shift the label distribution of the original data. Consequently, this phenomenon provides heuristics in practice, which indicates that excessively adding counterfactual augmentation data is inefficient.

\subsubsection{Addictive out-of-the-domain test}
\label{sec:Addictive out-of-the-domain test}
To further validate the out-of-domain generalization achieved by our data augmentation (DA) method, we use the model trained with original data (i.e., w/o DA in Section~\ref{sec:influence_of_augmentation_ratios}) as the baseline and the model trained with the combination of original data and 1-ratio CAD (i.e., Ratio 1 DA in Section~\ref{sec:influence_of_augmentation_ratios}) as the enhanced system. We test them on the ECB+ test corpus. The experimental results are presented in Table~\ref{tab:fcc_cross_ecb}.
\begin{table}[H]
    \centering
    \resizebox{1.0\columnwidth}{!}{
    \begin{tabular}{cccccc}
        \toprule
        \multirow{1}{*}{Methods} & \multicolumn{1}{c}{MUC} & \multicolumn{1}{c}{B$^3$}  & \multicolumn{1}{c}{CEAFe} & {LEA} & {CoNLL}\\
        \midrule
        Baseline System~\cite{held-etal-2021-focus}                   & 71.2 & 28.1 & 21.5 & 21.0 & 40.3 \\
        Enhanced System (ours) & \textbf{72.0} & \textbf{32.1} & \textbf{32.0} &\textbf{22.9} & \textbf{45.4}\\
        \bottomrule
    \end{tabular}
    }
    \caption{Cross-corpus evaluation by training the system on FCC but testing on ECB+.}
    \label{tab:fcc_cross_ecb}
\end{table}
The enhanced system consistently outperforms the baseline across all metrics, with a notable 5.1-point improvement in CoNLL F1. These results underscore the enhanced generalization achieved by our method in another out-of-domain scenario.

\subsubsection{Pairwise comparison via LLMs}
\label{sec:Pairwise comparison via LLMs}
\begin{table*}[t]
    \centering
    \small
    \begin{tabular}{ccccccc}
    \toprule
    Model (prompt type)                                     & R          & P          & F1         & \multicolumn{1}{l}{PR(\%)} & TComp(\%)     & Acc           \\ \midrule
    GPT-4 (zero-shot CoT)                                   & 93.7       & 57.5       & 71.3       & 55.8                       & 99.9       & 74.0            \\
    GPT-4 (few shots CoT)                                   & {\underline{95.5}} & 50.7       & 66.3       & {\underline{64.5}}                 & {\underline{100.0}}  & 66.7          \\ \midrule
    GPT-3.5-turbo (zero-shot CoT)                           & 67.0       & 53.3       & 59.4       & 43.0                       & 99.7       & 68.4          \\
    GPT-3.5-turbo (few shots CoT)                           & {\underline{78.6}} & 48.0       & {\underline{59.6}} & {\underline{56.2}}                 & 99.1       & 62.9          \\ \midrule
    Gemini-Pro (zero-shot CoT)                              & 62.5       & 48.0       & 54.3       & 41.8                       & 78.3       & 51.9          \\
    Gemini-Pro (few shots CoT)                              & {\underline {98.8}} & 43.8       & {\underline{60.7}} & {\underline{76.3}}                 & {\underline{92.3}} & {\underline{52.4}}    \\ \midrule
    Chat-Bison-001 (zero-shot CoT)                          & 90.1       & 44.7       & 59.7       & 68.4                       & 92.5       & 54.4          \\
    Chat-Bison-001 (few shots CoT)                          & {\underline{99.4}} & 35.9       & 52.7       & {\underline{94.9}}                 & {\underline{95.0}}   & 37.0            \\ \midrule
    Llama2-7b-chat (zero-shot CoT)                          & 94.0       & 34.5       & 50.5       & 93.3                       & 99.9       & 36.8          \\
    Llama2-7b-chat (few shots CoT)                          & {\underline{94.8}} & {\underline{34.8}} & {\underline{51.0}} & {\underline{93.6}}                 & 96.7       & 36.0            \\ \midrule
    \multicolumn{1}{l}{Our pipeline method (Roberta-Large)} & 90.0         & \textbf{83.7}       & \textbf{86.7}       & 36.8                       & \textbf{100.0}        & \textbf{90.6} \\ \bottomrule
    \end{tabular}
    \caption{Evaluation results for pairwise comparison. Better results than zero-shot CoT are underlined. }
    \label{tab:pairwise_comparision_LLMs}
\end{table*}
Comparing retrieved mention pairs is essentially treated as a binary classification problem, which is a significant intermediate step in current ECR systems~\cite{cattan2020streamlining,held-etal-2021-focus}. \citeauthor{yang2023supervised} demonstrates the strong ability of LLMs on several binary classification benchmarks~\cite{yang2023gluex}, such as SST-2, QNLI, etc. In this section, we explore the ability of LLMs to solve the event coreference comparison problem. 

The pairwise comparison dataset is consistent with the pairwise dataset used to evaluate our causally enhanced pipeline method, which is organized from the ECB+ test corpus. It comprises 6662 mention pairs, with 2282 coreferential (positive) examples and 4380 non-coreferential (negative) examples. 

As for prompts, we employ zero-shot and few-shot versions with the Chain-of-the-Thought (CoT)~\cite{wei2023chainofthought} strategy to query LLMs. To alleviate bias from few-shot examples, we select two positive examples and two negative examples from the training corpus as demonstrating examples. Such prompts are presented in Table~\ref{tab:zero-shot_CoT} and~\ref{tab:few-shot_CoT}. Within these prompts, LLMs are instructed to employ a 0-1 \textit{Coreferential score} to assess the degree of coreference between two events. We extract \textit{Coreferential results} directly from LLMs' responses as predicted labels for our evaluation.

Due to variations in the internal safety mechanisms and instruction-following capacity of LLMs, not all pairwise examples can be answered in the specified format, posing challenges for researchers in getting the concerned results from the responses of LLMs. Therefore, we propose the \textbf{T}ask \textbf{Comp}leteness (\textit{TComp}) metric to measure the completeness of LLMs for the pairwise comparison task:
\begin{equation}
    \label{completeness}
    \small
    TComp=\frac{N_{complete}}{N_{total}} 
\end{equation}
\footnotetext[4]{https://ai.google.dev/models/gemini}
\footnotetext[5]{https://ai.google.dev/models/palm}
\footnotetext[6]{https://huggingface.co/meta-llama/Llama-2-7b-chat}
where $N_{total}$ refers to the total number of pairwise examples fed to LLMs; $N_{complete}$ denotes the number of \textbf{\textit{completed examples}}, which are successfully answered by LLMs and easily parsed out by a pre-defined regular expression based on the prompt format. On top of that, we also measure the Recall (R), Precision (P), F1 and Positive prediction Rate (PR) on \textbf{\textit{completed examples}}. The accuracy (Acc) is measured on \textbf{\textit{total examples}}, which is the proportion of correctly predicted samples in the total pairwise dataset. It is used to measure the performance of the LLM. In our experiments, GPT-4, GPT-3.5-turbo, Gemini-Pro~\footnotemark[4]~\cite{geminiteam2023gemini}, Chat-Bison-001~\footnotemark[5]~\cite{anil2023palm}, and Llama2-7b-chat~\footnotemark[6]~\cite{touvron2023llama} are evaluated.

As shown in Table~\ref{tab:pairwise_comparision_LLMs}, the \textit{TComp} and accuracy of the GPT series models are significantly better than other LLMs, which means that GPTs essentially resolve the desired results for all pairwise comparison responses. Among them, GPT-4 demonstrates outstanding performance with a 74 Acc. However, there is still a significant deficiency compared to pipeline methods based on smaller models  (v.s. 90.6 Acc).

Gemini-Pro (zero-shot CoT) exhibits the lowest \textit{TComp} at 78.3\%. Adding few-shot demonstrating examples improves TComp to 92.3\%, but does not significantly enhance performance (zero-shot CoT 51.9 Acc v.s. few shots CoT 52.4 Acc). However, it's worth noting that Gemini-Pro demonstrates a robust ability to follow instructions. This is evidenced by the fact that all completed examples strictly adhere to our prompt requirements, while all incomplete examples are cases where the LLM refuses to respond due to its safety mechanism.    

GPT-4, GPT-3.5-turbo and Chat-Bison-001 show significant accuracy drops after introducing few-shot examples, with Chat-Bison-001 dropping by 17.4. This leads to decreased precision and a higher proportion of positive predictions, particularly severe in Chat-Bison-001, with 94.9\% of completed examples predicted as positive samples after the addition of few-shot examples.

For Llama2-7b-chat, \textit{TComp} is relatively good with above 95\%, but there is little change in performance after introducing few-shot examples. At the same time, this model also exhibits a strong positive bias, with over 93\% of completed examples predicted as positive under different prompt settings. Therefore, in future work, how to reduce the `positive bias' from LLMs when making coreference decisions is an interesting question to explore.

\subsection{Case study}
\subsubsection{Case study for LLM}
\label{sec:Case_study_for_LLM}
LLMs make three types of errors when addressing ECR: `Missing the golden mention', `Redundant mention prediction', and `Wrong mention prediction'. We provide examples and explanations for each error type.

\noindent \textbf{Missing the golden mention}. This error takes two types:
\begin{center}
    \small
    \fbox{\parbox{0.9\linewidth}{
    \textbf{Prompt:} Indonesia's West Papua province was [hit](\#) by a magnitude 6.1 [earthquake](\#) today...\\
    \textbf{Type 1:} Indonesia's West Papua province was [hit](\#) by a magnitude 6.1 [earthquake](\#) today...}
    }
\end{center}
In Type 1, the LLM has identified the golden mention marked with Markdown tags in the prompt but has failed to provide the required coreference result in the specified location.

\begin{center}
    \small
    \fbox{\parbox{0.9\linewidth}{
    \textbf{Prompt:} [Industry](\#) analysts [contacted](\#) by eWEEK generally [say](\#) they [believe](\#) that...\\
    \textbf{Type 2:} Industry analysts contacted by eWEEK generally say they believe that...}
    }
\end{center}
In Type 2, the LLM omits the golden mention marked in the prompt, while also failing to provide the coreference result.

Claude-2 misses 263 of 1780 golden mentions, 99\% being Type 1 errors, while GPT-4 misses 16, with 94\% being Type 2 errors. 

\noindent \textbf{Redundant mention prediction.} 
Redundant annotations sometimes occur in the LLM's response. For example:
\begin{center}
    \small
    \fbox{\parbox{0.9\linewidth}{
    \textbf{Prompt:} AMD [snaps up](\#) server upstart SeaMicro Much will be made of AMD [entering](\#)...\\
    \textbf{LLM response:} AMD [snaps up](\#190) server upstart SeaMicro Much will be \textcolor{red}{[made](\#191)} of AMD [entering](\#192)...}
    }
\end{center}
The number in the LLM response represents the predicted cluster-ID. The highlighted red portion indicates the redundant part generated by the LLM. We attribute this phenomenon to the LLM hallucination. 

Claude-2 generates 12 redundant mentions in total (11 mentions in Topic 41 and 1 mention in Topic 43), while GPT-4 makes no such error. As we could not identify corresponding golden clusters for these redundant predicted mentions, we excluded such instances when computing evaluation metrics.

\noindent \textbf{Wrong mention prediction.} 
This type of error is most common, indicating that certain mentions have not been properly clustered. For example:
\begin{center}
    \small
    \fbox{\parbox{0.9\linewidth}{
    \textbf{Ground truth:} Apple Inc. [took the wraps off](\#1) of [updated](\#2) iTunes and iLife software and [unveiled](\#1) a 17-inch MacBook Pro on Tuesday.\\
    \textbf{GPT-4 response:} Apple Inc. [took the wraps off](\#1) of [updated](\#2) iTunes and iLife software and [unveiled](\#3) a 17-inch MacBook Pro on Tuesday.}
    }
\end{center}
The ground truth presents two mention clusters: \{\textit{took the wraps off}, \textit{unveiled}\}, \{\textit{updated}\}, while the response of GPT-4 presents three mention clusters: \{\textit{took the warps off}\}, \{\textit{unveiled}\} and \{\textit{updated}\}. It means that GPT-4 assigns \textit{took the wraps off} and \textit{unveiled}, which should have been placed in the same cluster, to different clusters. 

Multiple coreference metrics in Table~\ref{tab:LLM_eval_zero_shot} evaluate the quality of clustering provided by LLMs, with GPT-4 demonstrating a significant advantage. 

\subsubsection{Case study for model interpretability}
\label{sec:Case_study_for_model_interpretability}
To further illustrate our causally enhanced system in terms of identifying and understanding event coreferential relationships, we present a mention-pair example from the ECB+ test set. The mention-pair example is inherently coreferential (Ground truth label: 1.0), but the baseline model predicts it as non-coreferential (Baseline predicted label: 0.0), resulting in a false negative (FN) example of the baseline model. 
\begin{center}
    \small
    \fbox{\parbox{0.9\linewidth}{
    \textbf{Mention-pair example (baseline FN)}\\
    1: ...\textbf{AMD} will \textit{\textbf{pay}} \$334 million for \textbf{SeaMicro}, including \$281 million in cash. ...\\
    2: ...\textbf{AMD} \textit{\textbf{shelled out}} \$334 million for the acquisition of \textbf{SeaMicro}. ...}}
\end{center}
The mention-pair example has a lexically divergent but semantically related trigger pair (\textit{\textbf{pay}}, \textit{\textbf{shelled out}}) and two participant pairs: (\textit{AMD}, \textit{AMD}), (\textit{SeaMicro}, \textit{SeaMicro}). Mentions in the above two texts both refer to the event `\textit{AMD paid for acquiring SeaMicro}'. To explore the model's sensitivity to `triggers lexical matching', we mechanically changed the first event trigger `\textbf{\textit{pay}}' to `\textbf{\textit{shelled out}}' as in modified example 1:
\begin{center}
    \small
    \fbox{\parbox{0.9\linewidth}{
    \textbf{Modified example 1 (trigger modification)}\\
    Text 1: ...\textbf{AMD} will \textit{\textbf{\textcolor{red}{shelled out}}} \$334 million for \textbf{SeaMicro}, including \$281 million in cash. ...\\
    Text 2: ...\textbf{AMD} \textit{\textbf{shelled out}} \$334 million for the acquisition of \textbf{SeaMicro}. ...}
    }
\end{center}
In modified example 1, the trigger pair becomes lexically similar. For the baseline model, we observe a significant prediction change: 0.17 (before the change) $\rightarrow$ 0.99 (after the change). This highlights the baseline model's high sensitivity to `triggers lexical matching', where a slight alteration in the trigger term's literal surface leads to a substantial impact on the prediction results in the same semantic context. In contrast, the causally enhanced model's prediction score remained unchanged: 0.99 (before the change) $\rightarrow$ 0.99 (after the change). This indicates that the enhanced model is more robust to semantic variations in triggers and is not easily influenced by the `triggers lexical matching' dilemma because it effectively alleviates the spurious associations in the decision-making process.

To explore the enhanced model's dependency on event-relevant arguments, which serve as rationales for ECR, we make small changes to participant arguments, such as `\textit{AMD}' $\rightarrow$ `\textit{Intel}' in text 1 of modified example 2:
\begin{center}
    \small
    \fbox{\parbox{0.9\linewidth}{
    \textbf{Modified example 2 (participant modification)}\\
    1: ...\textcolor{red}{\textbf{Intel}} will \textit{\textbf{pay}} \$334 million for \textbf{SeaMicro}, including \$281 million in cash. ...\\
    2: ...\textbf{AMD} \textit{\textbf{shelled out}} \$334 million for the acquisition of \textbf{SeaMicro}. ...}
    }
\end{center}
and `\textit{SeaMicro}' $\rightarrow$ `\textit{Cisco}' in text 1 of modified example 3: 
\begin{center}
    \small
    \fbox{\parbox{0.9\linewidth}{
    \textbf{Modified example 3 (participant modification)}\\
    1: ...\textbf{AMD} will \textit{\textbf{pay}} \$334 million for \textcolor{red}{\textbf{Cisco}}, including \$281 million in cash. ...\\
    2: ...\textbf{AMD} \textit{\textbf{shelled out}} \$334 million for the acquisition of \textbf{SeaMicro}. ...}
    }
\end{center}
Considering the changes in rationales, both modified example 2 and example 3 should be non-coreferential (Ground truth label: 0.0).

Consistent with our expectation, when making minimal changes to rationales, the enhanced system correctly predicates non-coreferential relationships. In modified example 2, the prediction of the enhanced system decreases from 0.99 (before the change) to 2.83e-05 (after the change). Similarly, in modified example 3, the prediction of the enhanced system decreases from 0.99 (before the change) to 5.82e-06 (after the change). This demonstrates the enhanced system's reliance on rationales, leading to more reasonable decisions based on causal associations, aligning with our motivation and SCM modelling.

Overall, modified example 1 illustrates that the enhanced model effectively avoids the misleading impact of `triggers lexical matching' on the results. Additionally, testing the enhanced model with modified examples 2 and 3, two counterfactual samples, reveals its ability to make correct judgments under a harsh condition. This indicates that the enhanced model can more effectively identify and understand event coreferential relationships based on pairwise semantic analysis.

\subsubsection{Case study for generalization test}
\label{sec:Case_study_for_generalization_test}
In the out-of-the-distribution setting, we compare the pairwise comparison results between the baseline system and our causally enhanced system for mention pairs retrieved from the test set. As shown in Table~\ref{tab:ood_pairwise_comparison}, the enhanced system significantly improves recall performance. This suggests that the enhanced system corrects a substantial number of false negative errors from the baseline, with 35.6\% of baseline false negatives (FNs) corrected to true positives (TPs). Among these resolved false negatives, 40\% of resolved FNs involve lexically divergent trigger pairs. By emphasizing counterpart arguments' coreference (i.e., rationales), the enhanced system enables the capture of clues of coreference in pairwise contexts, leading to the correct correction of these examples. In the main text, we have examined two types of error: `Ignore argument counterparts' and `Require contextual understanding'. Here, we provide examples and explanations for `Without domain expertise', `Lack of evidence' and `Annotation mistakes'.

\begin{table}[t]
    \centering
    \resizebox{1.0\columnwidth}{!}{
    \begin{tabular}{ccccc}
    \toprule
    Method          & R             & P             & F1             & Acc           \\ \midrule
    Baseline system~\cite{held-etal-2021-focus} & 36            & \textbf{76.5} & 48.96          & 54.8          \\
    Enhanced system & \textbf{54.7} & 71.9          & \textbf{62.13} & \textbf{59.8} \\ \bottomrule
    \end{tabular}}
    \caption{Pairwise comparison results for the OOD generalization test.}
    \label{tab:ood_pairwise_comparison}
\end{table}

\noindent \textbf{Without domain expertise.} 
This type of error can be attributed to the model's lack of domain-specific expertise. For example:
\begin{center}
    \small
    \fbox{\parbox{0.9\linewidth}{
    1. ...Didier Deschamps' side repeated the success on home soil at \textbf{\textit{France'98}} by a margin that hardly looked possible as Croatia stood toe-to-toe with the favourites for an hour...\\
    2. ... Pogba became the first Manchester United player to score in a World Cup final, and the first Premier League player to do so since Emmanuel Petit in \textbf{\textit{1998}}...}
    }
\end{center}
Here, the \textbf{\textit{France '98}} in text 1 and the \textbf{\textit{1998}} in text 2 refer to the `\textit{1998 FIFA World Cup France victory}'. However, without domain-specific knowledge, it is difficult to connect the two. Therefore, integrating more relevant knowledge during model training helps the model grasp the expertise and make accurate predictions. The enhanced system only resolved 3\% of such errors among the sampled 50 baseline false negatives (FNs) (see Figure~\ref{fig:error_analysis}). This poses a challenge for ECR systems trained in an OOD corpus to solve such errors, which may entail the enhanced system finding matching arguments from the pairwise context.

\noindent \textbf{Lack of the evidence.} 
\label{sec:lack_evidence_example}
In these examples, it is difficult to find evidence in the sentence pair to explain the ground truth. Perhaps we need to consider a longer context to find corresponding information. Considering an example from the false positives:
\begin{center}
    \small
    \fbox{\parbox{0.9\linewidth}{
    1. ...Of Eriksen's play so far in this World Cup, he said, "He has the capacity to do more." Read more about the \textbf{\textit{World Cup}}: No Ronaldo? No Messi? No problem: Nine names to know for the rest of the World Cup...\\
    2. ...Uruguay received more yellow cards (two) than they had in their previous four World Cup games combined (one). France have now scored with each of their last six shots on target at the \textbf{\textit{World Cup}}...}}
\end{center} 
According to the ground truth, these two mentions refer to the `FIFA World Cup 2018 tournament'. However, there is no direct evidence to explain why these two mentions are coreferential apart from the information provided by the trigger itself. To address these errors, more context may need to be added to the input. Our enhanced system guesses these examples correctly based on limited semantics.

\noindent \textbf{Annotation mistakes.} 
\label{sec:annotation mistakes}
In contrast to `Lack of the evidence', the context of these mention pairs provides sufficient information for humans to discern the ground truth mistakes. For example:
\begin{center}
    \small
    \fbox{\parbox{0.9\linewidth}{
    1. England holds its breath as World Cup \textbf{\textit{semi-final looms}}. Millions of expectant England fans will be glued to TV sets on Wednesday evening hoping the national team can reach the World Cup final for only the second time in their history...\\
    2. World Cup: England embraces Waistcoat Wednesday as football fever sweeps nation. As England gears up for tonight's \textbf{\textit{semi-final match}}, people across the country are aping Gareth Southgate and proudly wearing waistcoats to work...}}
\end{center} 
From the context, it is evident that the two mentions refer to `\textit{England's upcoming participation in the World Cup semi-final}'. However, according to the ground truth, these two events are not coreferential, which is nonsensical. Among all 100 samples we sampled, such instances accounted for 24\%. In future work, it is necessary to filter the annotation mistakes in the FCC dataset for a more plausible analysis.
\newpage

\begin{table*}[t]
    \resizebox{\linewidth}{!}{
        \begin{tabular}{l}
            \toprule
            \begin{tabular}[c]{@{}l@{}}
    \textbf{Demonstration:}\\Please perform a two-step task based on commonsense inference.\\Step1: Can you give me five similar expressions for the given word extracted from a sentence? Given word: \underline{`\textbf{\textit{fire}}'} from `\underline{A man has been }\\\underline{charged on suspicion of arson following a \textbf{\textit{fire}} that devastated a Somerset supermarket.}' Please show all expressions here.\\Step2: Try to use each expression in Step1 as the event head lemma to generate event mention not coreferential to the event induced by \\the event head lemma `\textbf{\textit{\underline{fire}}}' in the given text: \underline{`A man has been charged on suspicion of arson following a \textbf{\textit{fire}} that devastated a Somerset }\\\underline{supermarket.'} Noting that, the human participants, non-human participants, times and locations in generated event mention content should \\be not conferential to and different from those in the given text, keeping the sentence structure same as the given text as possible. \\Expressions: \textbf{\textit{blaze, inferno, conflagration, flames, combustion}}\\Event mentions:\\1. A woman has been charged with arson after a \textbf{\textit{blaze}} at the local library in Bristol.\\2. A teenager has been charged with arson after an \textbf{\textit{inferno}} at the shopping mall in Birmingham.\\3. A basketball athlete has been charged with arson after a \textbf{\textit{conflagration}} at the historic museum in York.\\...\\\textbf{Input:}\\Please perform a two-step task based on commonsense inference.\\Step1: Can you give me five similar expressions for the given word extracted from a sentence? Given word: `\textit{\textbf{\underline{died}}}' from\underline{ `Golden girl of screen }\\\underline{and pool, Esther Williams, has \textbf{\textit{died}} peacefully in her sleep aged 91.' }Please show all expressions here.\\Step2: Try to use each expression in Step1 as the event head lemma to generate event mention not coreferential to the event induced by the \\event head lemma `\textbf{\underline{\textit{died}}}' in the given text: \underline{`Golden girl of screen and pool, Esther Williams, has \textbf{\textit{died}} peacefully in her sleep aged 91.'} Noting \\that, the human participants, non-human participants, times and locations in generated event mention content should be not conferential to \\and different from those in the given text, keeping the sentence structure same as the given text as possible.~\\\textbf{LLM Response:}\\Expressions: \textbf{\textit{departed, expired, perished, left us, passed away}}\\1. The renowned musician Prince \textbf{\textit{departed}} from this world in his studio in Minneapolis at the age of 57.\\2. The legendary actor Marlon Brando \textbf{\textit{expired}} in his mansion in Los Angeles at the age of 80.\\3. The famous singer Whitney Houston \textbf{\textit{perished}} in her hotel room in New York at the age of 48.\\...\end{tabular}  \\
    \bottomrule
    \end{tabular}}
    \caption{The prompt about producing non-coreferential mention candidates for the input mention, where step 1 indicates the prompt operator $SYN$ and step 2 indicates the prompt operator $NCE$.}
    \label{tab:coref_syn_cf_prompt}
\end{table*}

\begin{table*}[t]
    \resizebox{\linewidth}{!}{
        \begin{tabular}{l}
            \toprule
            \begin{tabular}[c]{@{}l@{}}
    \textbf{Demonstration:}\\Please perform a two-step task based on commonsense inference. \\Step1: Can you give me five similiar expressions for the given word extracted from the sentence? Given word:\underline{`\textbf{\textit{free throw}}' }from \underline{`McDermott}\\\underline{broke Rodney Buford' s school scoring record of 2,116 points with a \textbf{\textit{free throw}} with 4:43 to play in the first half.'} Please show all expressions here.\\Step2: Try to use each expression in Step1 as the event head lemma to generate event mention coreferential to the event induced by the event \\head lemma \textbf{\textit{\underline{`free throw'}}} in the given text: \underline{`McDermott broke Rodney Buford's school scoring record of 2,116 points with a \textbf{\textit{free throw}} with 4:43 to }\\\underline{play in the first half.'} Noting that, the human participants, non-human participants, times and locations in generated event mention content should \\be coreferential to those in the given text, but the sentence structure can be different from the given text.\\Expressions: \textbf{\textit{basketball shot, scoring toss, uncontested shot, charity toss, foul shot}}\\Event mentions:\\1. McDermott broke Rodney Buford's school scoring record of 2,116 points with a \textbf{\textit{basketball shot}} at 4:43 remaining in the first half.\\2. McDermott surpassed Rodney Buford's school scoring record of 2,116 points with a \textbf{\textit{scoring toss}} in the first half with 4:43 left on the clock.\\3. McDermott set a new school scoring record of 2,116 points with an \textbf{\textit{uncontested shot}} during the first half with 4:43 remaining.\\...\\\textbf{Input:}\\Please perform a two-step task based on commonsense inference.\\Step1: Can you give me five similar expressions for the given word extracted from a sentence? Given word: \underline{\textbf{\textit{`protect'}}} from `\underline{"Microsoft has released a }\\\underline{security update for Internet Explorer that will help \textbf{\textit{protect}} its customers from malicious attacks", said the company in a statement.'} Please show all \\expressions here.\\Step2: Try to use each expression in Step1 as the event head lemma to generate event mention not coreferential to the event induced by the event head \\lemma \textbf{\textit{`\underline{protect}'}} in the given text:`\underline{"Microsoft has released a security update for Internet Explorer that will help~\textbf{\textbf{\textit{protect}}}~its customers from malicious attacks'', }\\\underline{said the company in a statement.'}~Noting that, the human participants, non-human participants, times and locations in generated event mention content should \\be coreferential to those in the given text, but the sentence structure can be different from the given text.\\\textbf{LLM Response:}\\Expressions: \textbf{\textit{secured,~fortifying, defend, safeguard,~shield}}\\1. A statement from Microsoft confirms that the company has \textbf{\textit{secured}} its customers from malicious attacks by releasing a security update for Internet Explorer.~\\2. Microsoft has released a security update for Internet Explorer, \textbf{\textit{fortifying}} its customers against malicious attacks, according to a recent statement.\\3. Microsoft has released a security update for Internet Explorer in order to \textbf{\textit{defend}} its customers from malicious attacks, according to a statement.\\...\end{tabular}  \\
    \bottomrule
    \end{tabular}}
    \caption{The prompt about producing coreferential mention candidates for the input mention, where step 1 indicates the prompt operator $SYN$ and step 2 indicates the prompt operator $CE$.}
    \label{tab:non_coref_syn_cf_prompt}
\end{table*}
\newpage

\begin{table*}[t]
    \resizebox{\linewidth}{!}{
        \begin{tabular}{l}
            \toprule
            \begin{tabular}[c]{@{}l@{}}
    \textbf{Demonstration:}\\We have a snippet of text: \underline{`Indianapolis Colts clinch playoff berth with win over Kansas City Chiefs December 23, 2012. Going into week 16, the Indianapolis }\\\underline{Colts controlled their own destiny of making it to post-season play. The Colts could clinch a playoff berth with a win over the Kansas City Chiefs or a Pittsburgh }\\\underline{Steelers loss. As they have done all season, the Colts refused to let their fate be decided by anyone other than themselves. The young team fought hard to }\\\underline{defeat the Chiefs in another fourth quarter victory, 20-13. Although they started the game with a three-and-out, the Colts were able to light up the scoreboard first.' }\\The text can be divided into prefix, mention and suffix as following:\\Prefix: \underline{`Indianapolis Colts clinch playoff berth with win over Kansas City Chiefs December 23, 2012. Going into week 16, the Indianapolis Colts controlled their }\\\underline{own destiny of making it to post-season play.' }\\Mention:\underline{ `The Colts could clinch a playoff berth with a win over the Kansas City Chiefs or a Pittsburgh Steelers loss.' }\\Suffix: \underline{`As they have done all season, the Colts refused to let their fate be decided by anyone other than themselves. The young team fought hard to defeat the }\\\underline{Chiefs in another fourth quarter victory, 20-13. Although they started the game with a three-and-out, the Colts were able to light up the scoreboard first.'}\\Can you paraphrase Prefix and Suffix in five different ways, where human participants, non-human participants, times, locations and actions in generated examples \\are conferential to those in the original one? \\Prefix: \\1. The Indianapolis Colts secured a spot in the playoffs by defeating the Kansas City Chiefs on December 23, 2012, in week 16 of the season. The Colts had control \\of their own destiny and could have also clinched a playoff berth with a Pittsburgh Steelers loss.\\2. On December 23, 2012, the Indianapolis Colts earned a playoff spot by winning against the Kansas City Chiefs in week 16. The Colts had the power to determine \\their own fate and could have also secured a playoff berth if the Pittsburgh Steelers lost.\\...\\Suffix:\\1. Throughout the season, the Indianapolis Colts refused to let anyone else decide their fate. In another fourth-quarter victory, the young team fought hard to defeat \\the Kansas City Chiefs with a score of 20-13, despite starting the game with a three-and-out.\\2. The Indianapolis Colts demonstrated their determination to control their own destiny throughout the season. They fought hard to secure another fourth-quarter \\victory against the Kansas City Chiefs, winning 20-13, despite starting the game with a three-and-out.\\...\\\textbf{\textbf{Input:}}\\We have a snippet of text: \underline{`Microsoft has said that an emergency security update has fixed a flaw in Internet Explorer that left millions of computers vulnerable }\\\underline{to hacking and hijack. The software patch, which was released last night, has closed a loophole that has seen the computers of at least two million users hacked }\\\underline{by cyber criminals. The problem related to a "zero day" flaw that tricked people into visiting an infected website, enabling hackers to gain access to online banking }\\\underline{passwords and e-shopping logon details. "Microsoft has released a security update for Internet Explorer that will help protect its customers from malicious attacks", }\\\underline{said the company in a statement." Like a vaccine developed to fight a virus, this "security update" will protect computers only if it is installed." Computers that are }\\\underline{set to automatically update and install software will already be protected by the patch. Those users who manage their own updates are advised to download the fix }\\\underline{as soon as possible from the Microsoft website.' }\\The text can be divided into prefix, mention and suffix as following:\\Prefix: \underline{`Microsoft has said that an emergency security update has fixed a flaw in Internet Explorer that left millions of computers vulnerable to hacking and hijack. }\\\underline{The software patch, which was released last night, has closed a loophole that has seen the computers of at least two million users hacked by cyber criminals. The }\\\underline{problem related to a "zero day" flaw that tricked people into visiting an infected website, enabling hackers to gain access to online banking passwords and e-shopping }\\\underline{logon details.' }\\Mention: "\underline{'Microsoft has released a security update for Internet Explorer that will help protect its customers from malicious attacks', said the company in a statement." }\\Suffix:\underline{ `Like a vaccine developed to fight a virus, this "security update" will protect computers only if it is installed." Computers that are set to automatically update }\\\underline{and install software will already be protected by the patch. Those users who manage their own updates are advised to download the fix as soon as possible from the }\\\underline{Microsoft website.'}\\Can you paraphrase Prefix and Suffix in five different ways, where human participants, non-human participants, times, locations and actions in generated examples are \\conferential to those in the original one?\\\textbf{LLM Response:}\\Prefix: \\1. Microsoft has addressed a flaw in Internet Explorer that left millions of computers vulnerable to hacking and hijack by releasing an emergency security update. \\The software patch, which was released last night, has closed a loophole that allowed cyber criminals to hack into the computers of at least two million users by exploiting \\a "zero day" flaw that tricked people into visiting an infected website.\\2.An emergency security update from Microsoft has fixed a flaw in Internet Explorer that left millions of computers vulnerable to hacking and hijack. The software patch, \\which was released last night, has closed a loophole that allowed cyber criminals to hack into the computers of at least two million users by exploiting a "zero day" flaw \\that enabled them to gain access to online banking passwords and e-shopping logon details.\\...\\Suffix:\\1. Microsoft has likened the security update for Internet Explorer to a vaccine that helps protect computers from malicious attacks. The update will only be effective if it \\is installed, and computers set to automatically update and install software will already be protected by the patch. Users who manage their own updates are advised to \\download the fix from the Microsoft website as soon as possible. \\2. The security update for Internet Explorer released by Microsoft is similar to a vaccine that helps protect computers from malicious attacks. However, the update will \\only be effective if it is installed. Computers set to automatically update and install software will already be protected by the patch, but those who manage their own \\updates should download the fix from the Microsoft website as soon as possible.\end{tabular}  \\
    \bottomrule
    \end{tabular}}
    \caption{The prompt about paraphrasing the discourse context (i.e., prompt operator $PARA$).}
    \label{tab:para}
\end{table*}
\newpage

\begin{table*}[t]
    \resizebox{\linewidth}{!}{
        \begin{tabular}{l}
            \toprule
            \begin{tabular}[c]{@{}l@{}}
    \textbf{Demonstration:}\\
    We have a template sentence: \underline{`A publicist says Tara Reid has checked herself into rehab.'} Please generate three Prefixes and Suffixes \\for the template sentence. Prefix content should be about what typically happens before the event head lemma \underline{`checked herself'} in \\the given template sentence, while Suffix content should be about what typically happens after the event head lemma \underline{`checked herself'} \\in the given template sentence. Note: Each generated Prefix or Suffix contains three sentences.\\
    Prefixes:\\
    1. `After a series of public appearances where she appeared to be under the influence, rumors began to circulate that Tara Reid was \\struggling with addiction. Friends and family members reportedly urged her to seek help and get treatment before things got worse.'\\
    2. `Tara Reid often faces mounting pressure from friends, family, and inner demons. The weight of her addiction or mental health \\challenges becomes increasingly burdensome. Seeking relief and stability, she reaches a breaking point where seeking professional \\help is no longer just an option, but a necessity.'\\...\\
    Suffixes:\\
    1. `Reid's representatives have confirmed that she is taking her recovery seriously and is committed to staying in rehab for as long as\\ necessary. She has also expressed gratitude for the support she has received from fans and loved ones during this difficult time. It is\\ hoped that with the help of professionals, she will be able to overcome her addiction and move forward in a positive direction.'\\
    2. `Tara Reid begins a transformative journey towards healing after checking herself into rehab. She commits herself to a comprehensive\\ treatment plan tailored to her specific needs. With dedication and the support of professionals, she embarks on a path of self-discovery, \\growth, and sobriety.'\\
    ...\\
    
    \textbf{Input:}\\
    We have a template sentence: \underline{`After stealing a valuable painting from the museum, the thief managed to escape through the back door.'} \\Please generate three Prefixes and Suffixes for the template sentence. Prefix content should be about what typically happens before the \\event head lemma \underline{`escape'} in the given template sentence, while Suffix content should be about what typically happens after the event \\head lemma \underline{`escape'} in the given template sentence. Note: Each generated Prefix or Suffix contains three sentences.\\
    
    \textbf{LLM Response:}\\
    Prefixes:\\
    1. `The thief carefully scoped out the museum for weeks, studying the security protocols and identifying the perfect time to strike.\\They communicated with a network of fellow criminals, planning the heist down to the smallest detail. Once the valuable painting caught\\ their eye, they knew it was time to put their plan into action.'\\
    2. `Feeling the thrill of the adrenaline pumping through their veins, the thief meticulously studied the layout of the museum, looking for\\ weaknesses in the security system. They spent days plotting their escape route, ensuring that every step was calculated and foolproof.\\The moment they laid eyes on the valuable painting, they knew this was their chance to execute the perfect crime.'\\
    ...\\
    Suffixes:\\
    1. `The thief disappeared into the night, leaving authorities baffled and determined to bring them to justice. They carefully weaved through\\ a labyrinth of narrow alleyways, evading capture at every turn. With the stolen painting securely hidden, they planned their next move, \\aware that the chase was far from over.'\\
    2. `After successfully evading the museum security, the thief meticulously covered their tracks, leaving no evidence behind. They vanished into \\the underworld, becoming a legend among thieves. With the valuable painting in their possession, they now had the difficult task of finding a \\buyer who would be discreet and willing to pay a handsome sum.'\\
    ...\\
    \end{tabular}  \\
    \bottomrule
    \end{tabular}}
    \caption{The prompt about producing temporal commonsense prefixes and suffixes for a given event-mention sentence with its trigger. (i.e., prompt operator $TC$). 
    This prompt is inspired by~\citeauthor{ravi2023happens}'s work.}
    \label{tab:tc_prompt}
\end{table*}
\newpage

\begin{table*}[t]
    \resizebox{\linewidth}{!}{
        \begin{tabular}{l}
            \toprule
            \begin{tabular}[c]{@{}l@{}}
            \textbf{ORI}\\
            \textless{}s\textgreater{}Esther Williams-Esther Williams: "Million Dollar Mermaid" Swimmer And Actress Dies Aged 91. 07 June 2013. \underline{Golden girl of }\\\underline{screen and pool, Esther Williams, has \textbf{\textit{died}} peacefully in her sleep aged 91.} It has been confirmed by publicist Harlan Boll that \\the 1940s Hollywood actress and record-setting swimmer, Esther Williams, died on 6th June in her sleep. Stunning Williams \\quickly achieved movie success and pin-up status, due to her swimmer's physique being regularly snapped in bathing suits.\textless{}\textbackslash{}s\textgreater{}\\\textless{}s\textgreater{}Esther Williams, Olympic swimmer turned actress and pinup girl, dies at 91. Esther Williams, the swimming champion turned \\actress who starred in glittering and aquatic Technicolor musicals of the 1940s and 1950s, has died. She was 91. \underline{Williams \textit{\textbf{died }}}\\\underline{early Thursday in her sleep, according to her longtime publicist Harlan Boll.} Williams in a bathing suit became a favorite pinup of \\GI's in World War II , and her popularity continued afterward .\textless{}\textbackslash{}s\textgreater{}\\
            \textbf{CAD}\\
            \textless{}s\textgreater{}Esther Williams-Esther Williams: "Million Dollar Mermaid" Swimmer And Actress Dies Aged 91. 07 June 2013. \underline{The renowned }\\\underline{musician Prince \textbf{\textit{departed}} from this world in his studio in Minneapolis at the age of 57. }It has been confirmed by publicist Harlan \\Boll that the 1940s Hollywood actress and record-setting swimmer, Esther Williams, died on 6th June in her sleep. Stunning Williams \\quickly achieved movie success and pin-up status, due to her swimmer's physique being regularly snapped in bathing suits.\textless{}\textbackslash{}s\textgreater{}\\\textless{}s\textgreater{}Esther Williams, Olympic swimmer turned actress and pinup girl, dies at 91. Esther Williams, the swimming champion turned \\actress who starred in glittering and aquatic Technicolor musicals of the 1940s and 1950s, has died. She was 91. \underline{Williams \textit{\textbf{died }}}\\\underline{early Thursday in her sleep, according to her longtime publicist Harlan Boll.} Williams in a bathing suit became a favorite pinup of \\GI's in World War II , and her popularity continued afterward .\textless{}\textbackslash{}s\textgreater{}\\
            \textbf{TIA}\\
            \textless{}s\textgreater{}Esther Williams-Esther Williams: "Million Dollar Mermaid" Swimmer And Actress Dies Aged 91. 07 June 2013. \underline{The renowned }\\\underline{musician Prince \textbf{\textit{died}} from this world in his studio in Minneapolis at the age of 57.} It has been confirmed by publicist Harlan Boll \\that the 1940s Hollywood actress and record-setting swimmer, Esther Williams, died on 6th June in her sleep. Stunning Williams \\quickly achieved movie success and pin-up status, due to her swimmer's physique being regularly snapped in bathing suits.\textless{}\textbackslash{}s\textgreater{}\\
            \textless{}s\textgreater{}Esther Williams, Olympic swimmer turned actress and pinup girl, dies at 91. Esther Williams, the swimming champion turned \\actress who starred in glittering and aquatic Technicolor musicals of the 1940s and 1950s, has died. She was 91. \underline{Williams \textit{\textbf{died }}}\\\underline{early Thursday in her sleep, according to her longtime publicist Harlan Boll.} Williams in a bathing suit became a favorite pinup of \\GI's in World War II , and her popularity continued afterward .\textless{}\textbackslash{}s\textgreater{}\\
            \textbf{CIA}\\
            \textless{}s\textgreater{}On June 6, 2013, Esther Williams, the iconic Hollywood actress and swimmer, passed away peacefully in her sleep at the age of 91.\\ \underline{The renowned}  \underline{musician Prince \textbf{\textit{departed}} from this world in his studio in Minneapolis at the age of 57. } Esther Williams, the Hollywood \\actress and record-breaking swimmer, passed away on June 6, 2013, at the age of 91. Williams' beauty and athleticism made her a \\beloved figure in both the film industry and the world of swimming.\textless{}\textbackslash{}s\textgreater{}\\\textless{}s\textgreater{}Esther Williams, the Olympic swimmer who later became an actress and pinup girl, passed away at the age of 91. She was known \\ for her dazzling performances in Technicolor musicals during the 1940s and 1950s. \underline{Williams \textbf{\textit{died}} early Thursday in her}\\ \underline{sleep, according to her longtime publicist Harlan Boll. }Esther Williams' beauty and talent made her a favorite pinup of GI's during \\World War II, and her legacy continued to captivate audiences for years to come.\textless{}\textbackslash{}s\textgreater{}\\
            \textbf{TAD}\\
            \textless{}s\textgreater{}In the weeks leading up to his tragic departure, rumors began to circulate about Prince's declining health. Concerned fans \\and loved ones expressed their worries, hoping he would seek proper medical attention and take care of himself. \underline{The renowned }\\\underline{musician Prince \textbf{\textit{departed}} from this world in his studio in Minneapolis at the age of 57. } Tributes poured in, celebrating his iconic \\career and the impact he had on the world of music. His immense talent and legacy continue to inspire new generations of musicians\\ and fans alike.\textless{}\textbackslash{}s\textgreater{}\\\textless{}s\textgreater{}Esther Williams, like many people in the public eye, faced her fair share of personal struggles and demons. She had battled with\\ addictions and mental health issues throughout her life, which had impacted her overall health. Despite attempts to seek treatment and\\ find stability, her health continued to deteriorate, ultimately leading to her untimely death. \underline{Williams \textit{\textbf{died }}} \underline{early Thursday in her sleep, }\\\underline{according to her longtime publicist Harlan Boll.} Following her passing, Williams' publicist Harlan Boll released a statement expressing\\ his condolences and sharing the sadness of her death. Boll highlighted the impact Williams had on the entertainment industry and the void\\ her absence would leave. He also mentioned plans for a memorial service to honor her memory and celebrate her life\textless{}\textbackslash{}s\textgreater{}\\

            \end{tabular}  \\
    \bottomrule
    \end{tabular}}
    \caption{A coreferential original data (ORI) with its Counterfactual Augmented Data (CAD), Trigger Intervention Ablation (TIA) Data, Context Intervention Ablation (CIA) Data and Temporal commonsense Augmented Data (TAD). Mention sentences are underlined, with bold trigger terms.}
    \label{tab:cad_tia_cia_example}
\end{table*}

\begin{table*}[t]
    \resizebox{\linewidth}{!}{
        \begin{tabular}{l}
            \toprule
            \begin{tabular}[c]{@{}l@{}}
    \textbf{ORI}\\
    \textless{}s\textgreater{}Microsoft Rushes Emergency Fix To Address Internet Explorer Attacks. September 17, 2013 4:16 PM ET. \underline{Microsoft has rushed out }\\\underline{a temporary fix to \textbf{\textit{address}} ongoing attacks targeting an Internet Explorer zero-day vulnerability. }The software giant said the Fix-It \\temporary workaround should be effective in preventing a successful attack. The company said the vulnerability impacts all currently \\supported versions of the browser, but attacks have been limited to users of Internet Explorer 8 and Internet Explorer 9. "On completion \\of this investigation, Microsoft will take the appropriate action to protect our customers, which may include providing a solution through \\our monthly security update release process , or an out-of-cycle security update, depending on customer needs, " the company said in \\a security advisory issued Tuesday.\textless{}\textbackslash{}s\textgreater{}\\\textless{}s\textgreater{}Microsoft has said that an emergency security update has fixed a flaw in Internet Explorer that left millions of computers vulnerable \\to hacking and hijack. The software patch, which was released last night, has closed a loophole that has seen the computers of at least \\two million users hacked by cyber criminals. The problem related to a \`{}\`{}zero day'' flaw that tricked people into visiting an infected website, \\enabling hackers to gain access to online banking passwords and e-shopping logon details. \underline{\`{}\`{}Microsoft has released a security update for }\\\underline{Internet Explorer that will help \textit{\textbf{protect}} its customers from malicious attacks,'' said the company in a statement.} \`{}\`{}Like a vaccine developed \\to fight a virus , this 'security update' will protect computers only if it is installed. ''Computers that are set to automatically update and install \\software will already be protected by the patch. Those users who manage their own updates are advised to download the fix as soon as \\possible from the Microsoft website.\textless{}\textbackslash{}s\textgreater{}\\
    \textbf{CAD}\\
    \textless{}s\textgreater{}An emergency security update from Microsoft has fixed a flaw in Internet Explorer that left millions of computers vulnerable to hacking\\ and hijack. The software patch, which was released last night, has closed a loophole that allowed cyber criminals to hack into the computers\\of at least two million users by exploiting a "zero day" flaw.\underline{ A statement from Microsoft confirms that the company has \textbf{\textit{secured}} its customers }\\\underline{from malicious attacks by releasing a security update for Internet Explorer.} ' The security update for Internet Explorer released by Microsoft \\is like a vaccine that protects computers from malicious attacks. However, the update will only be effective if it is installed. Computers set to \\automatically update and install software will already be protected by the patch, but those who manage their own updates should download \\the fix from the Microsoft website without delay.\textless{}\textbackslash{}s\textgreater{}\\\textless{}s\textgreater{}Microsoft has said that an emergency security update has fixed a flaw in Internet Explorer that left millions of computers vulnerable\\to hacking and hijack. The software patch, which was released last night, has closed a loophole that has seen the computers of at least\\two million users hacked by cyber criminals. The problem related to a \`{}\`{}zero day'' flaw that tricked people into visiting an infected website,\\enabling hackers to gain access to online banking passwords and e-shopping logon details.~\underline{\`{}\`{}Microsoft has released a security update for}\\\underline{Internet Explorer that will help~\textit{\textbf{\textbf{protect}}}~its customers from malicious attacks,'' said the company in a statement.}~\`{}\`{}Like a vaccine developed\\to fight a virus , this 'security update' will protect computers only if it is installed. ''Computers that are set to automatically update and install\\software will already be protected by the patch. Those users who manage their own updates are advised to download the fix as soon as\\possible from the Microsoft website.\textless{}\textbackslash{}s\textgreater{}\\\textbf{TIA}\\
    \textless{}s\textgreater{}An emergency security update from Microsoft has fixed a flaw in Internet Explorer that left millions of computers vulnerable to hacking\\ and hijack. The software patch, which was released last night, has closed a loophole that allowed cyber criminals to hack into the computers\\ of at least two million users by exploiting a "zero day" flaw.~\underline{A statement from Microsoft confirms that the company has \textbf{\textit{protected}}~its customers}\\\underline{from malicious attacks by releasing a security update for Internet Explorer.}~' The security update for Internet Explorer released by Microsoft \\is like a vaccine that protects computers from malicious attacks. However, the update will only be effective if it is installed. Computers set to \\automatically update and install software will already be protected by the patch, but those who manage their own updates should download \\the fix from the Microsoft website without delay.\textless{}\textbackslash{}s\textgreater{}\\\textless{}s\textgreater{}Microsoft has said that an emergency security update has fixed a flaw in Internet Explorer that left millions of computers vulnerable\\to~hacking and hijack. The software patch, which was released last night, has closed a loophole that has seen the computers of at least\\two million users hacked by cyber criminals. The problem related to a \`{}\`{}zero day'' flaw that tricked people into visiting an infected website,\\enabling hackers to gain access to online banking passwords and e-shopping logon details.~\underline{\`{}\`{}Microsoft has released a security update for}\\ \underline{Internet Explorer that will help~\textit{\textbf{\textbf{\textbf{\textbf{protect}}}}}~its customers from malicious attacks,'' said the company in a statement.}~\`{}\`{}Like a vaccine developed\\to fight a virus, this 'security update' will protect computers only if it is installed. ''Computers that are set to automatically update and install\\software will already be protected by the patch. Those users who manage their own updates are advised to download the fix as soon as\\possible from the Microsoft website.\textless{}\textbackslash{}s\textgreater{}\\\textbf{CIA}\\
    \textless{}s\textgreater{}In response to the Internet Explorer attacks, Microsoft has rapidly provided an emergency fix on September 17, 2013, at 4:16 PM ET. \\\underline{A statement from Microsoft confirms that the company has \textbf{\textit{secured}} its customers from malicious attacks by releasing a security update for }\\\underline{Internet Explorer. }' According to Microsoft, the temporary Fix-It workaround is capable of preventing a successful attack on all currently supported \\versions of the browser. However, the attacks have only affected Internet Explorer 8 and 9 users.\textless{}\textbackslash{}s\textgreater{}\\\textless{}s\textgreater{}An emergency security update from Microsoft has fixed a flaw in Internet Explorer that left millions of computers vulnerable to hacking and \\hijack. The software patch, which was released last night, has closed a loophole that allowed cyber criminals to hack into the computers of at \\least two million users by exploiting a "zero day" flaw \`{}\underline{\`{}Microsoft has released a security update for Internet Explorer that will help \textbf{\textit{protect}} its }\\\underline{customers from malicious attacks,'' said the company in a statement.} The security update for Internet Explorer released by Microsoft is like a \\vaccine that protects computers from malicious attacks. However, the update will only be effective if it is installed. Computers set to automatically \\update and install software will already be protected by the patch, but those who manage their own updates should download the fix from the \\Microsoft website without delay.\textless{}\textbackslash{}s\textgreater{}\\
    \textbf{TAD}\\
    \textless{}s\textgreater{}Leading up to the release of the new security update, Microsoft has been monitoring user reports and conducting extensive testing to identify \\and understand the nature of the glitch in Internet Explorer. They have been working diligently behind the scenes to develop an effective solution.\\\underline{ A statement from Microsoft confirms that the company has \textbf{\textit{secured}} its customers from malicious attacks by releasing a security update for} \\\underline{Internet Explorer.'} Now that the security update has been released, Microsoft urges all users of Internet Explorer, including those on Windows 7\\ with IE11, to promptly install the patch to fix the glitch. They emphasize the importance of keeping browsers up to date to ensure optimal security\\ and protection against potential cyber threats. \textless{}\textbackslash{}s\textgreater{}\\\textless{}s\textgreater{}As part of their proactive approach to cybersecurity, Microsoft regularly conducts audits and vulnerability assessments of Internet Explorer.\\ During one such routine assessment, the company discovered a glitch that could potentially exploit users' systems. This prompted the swift develop-\\-ment of a security update to mend the vulnerability.~\underline{\`{}\`{}Microsoft has released a security update for}\underline{Internet Explorer that will help~\textit{\textbf{\textbf{protect}}} its custo-}\\\underline{-mers from malicious attacks,'' said the company in a statement.} With the successful release of the security update, Microsoft assures users that\\ their browsing experience with Internet Explorer will now be more secure and free from the glitch. They encourage users to reach out to their support\\ team if they encounter any further issues or require additional assistance.\textless{}\textbackslash{}s\textgreater{}\\
    \end{tabular}\\
    \bottomrule
    \end{tabular}}
    \caption{A non-coreferential original data (ORI) with its Counterfactual Augmented Data (CAD), Trigger Intervention Ablation (TIA) Data, Context Intervention Ablation (CIA) Data and Temporal commonsense Augmented Data(TAD).}
    \label{tab:non_coref_cad_tia_cia_example}
\end{table*}

\begin{table*}[t]
    \resizebox{\linewidth}{!}{
        \begin{tabular}{l}
            \toprule
            \begin{tabular}[c]{@{}l@{}}
                \textbf{Input:} \\
                We have two texts, each of which contains an event mention that we are interested in, with its head lemma marked with *bold*. Please determine whether these two events are coreferential.\\ Events co-refer only when human participants, non-human participants, locations, times, and actions where events occur co-refer. Your answer should satisfy the following format strictly:\\
                \textit{Rationales:} Let's think step by step and show the reason here.\\
                \textit{Coreferential results:} Answer it with only the word `Coreferential' or  `Non-Coreferential'.\\
                \textit{Coreferential score:} Assign a score between 0 and 1 to measure the degree of co-reference between the two events. A higher numerical value indicates a higher degree of co-reference.\\
                \\
                \textit{text1:} Maybe that's why Intel and Nvidia (NVDA), the companies' two biggest rivals, found a lot to smile about on July 24, the day the deal was announced. ``I thought it was just impossible\\ to get a gift like this,'' crowed Nvidia CEO Jen-Hsun Huang, in an interview with BusinessWeek.com. ATI is ``basically throwing in the towel, leaving us as the only stand-alone (graphics\\ chip) company in the world.'' Intel, the world's largest graphics chipmaker, declined to comment specifically on the \textbf{\textit{*deal*}}. But company execs have noted Intel continues to improve its own\\ graphics chipsets and that it expects to deliver platforms of products, like Centrino, that AMD can not easily match. Investors weren't too sure how happy the AMD and ATI family will turn\\ out to be, either. AMD shares fell almost 5\% on July 24, to \$17.40, extending recent declines fueled by concern AMD will be the biggest loser amid an escalating PC industry price war. \\
                \textit{text2:} Addressing a deficiency of its own and further upping the competitive tension between it and chief rival Intel (Nasdaq: INTC-news-people), Advanced Micro Devices (nyse: AMD-news\\-people) intends to acquire graphics chipmaker ATI Technologies (nasdaq: ATYT-news-people ) in a \$5.4 billion deal. Is Intel below \$19 too cheap to last or still above what the chipmaker's\\ worth? The question for many is where that leaves Nvidia Corp. (nasdaq: NVDA - news - people), now the sole independent graphics chip maker addressing each of the computing, mobile\\  communications and display markets. We think not too much worse off than before the \textbf{\textit{*deal*}} and believe shares in the company remain a compelling way to play growth in gaming- and media-\\ oriented computing. Our view of the Advanced Micro Devices/ATI Technologies deal is that it's the last major move AMD needed to make to match Intel's breadth in the processor market (now\\ that Intel has shed its communications processor business). Intel has had for some time the ability to integrate graphics capabilities into its offerings for PC makers, resources for which AMD\\ has had to rely on other folks, including both ATI and Nvidia. The AMD/ATI combination now trumps Intel's strategy by adding the capacity to package higher-end chips for gaming and\\ complex graphic manipulation\\
                \textit{Rationales:} \\
                \textit{Coreferential results:}\\
                \textit{Coreferential score: } \\
                \textbf{GPT-4 (Easy to parse \textit{Coreferential results}, True)}:\\
                \textit{Rationales:} Both texts are talking about the same event, which is the acquisition of ATI Technologies by Advanced Micro Devices (AMD). The human participants involved are the same (AMD, \\ATI Technologies, Intel, Nvidia), the non-human participants are the\\ same (the companies, the graphics chips), the location is not specified but can be assumed to be the same, and the time is the same (the deal was announced on July 24). The action is also the same,\\ which is the acquisition or deal.\\
                \textit{Coreferential results:} Coreferential\\
                \textit{Coreferential score:} 1\\
                \textbf{GPT-3.5-turbo (Easy to parse \textit{Coreferential results}, False)}:\\
                \textit{Rationales:}\\
                1. In text1, there is a mention of a deal being announced on July 24. In text2, there is also a mention of a deal. We need to determine if these two mentions refer to the same event.\\
                2. In text1, the deal is described as being between Intel and Nvidia, while in text2, the deal is described as being between Advanced Micro Devices (AMD) and ATI Technologies. We need to deter-\\-mine if these are referring to the same deal.\\
                3. In text1, the deal is mentioned in the context of Intel and Nvidia being rivals, while in text2, the deal is mentioned in the context of AMD and Intel being rivals. We need to determine if these rival-\\-ries are related to the same deal.\\
                \textit{Coreferential results:} Non-Coreferential\\
                \textit{Coreferential score:} 0.2\\
                \textbf{Gemini-Pro (Easy to parse \textit{Coreferential results}, True)}:\\
                \textit{Rationales:} The two events mentioned in the texts are both about the deal between Advanced Micro Devices (AMD) and ATI Technologies. The events share the same participants (AMD and ATI), the\\ same action (deal), and the same time (July 24).\\
                \textit{Coreferential results:} Coreferential\\
               \textit{Coreferential score:} 1\\
                \textbf{Chat-Bison-001 (Hard to parse \textit{Coreferential results}, False)}:\\
                The two events in the texts are "AMD acquires ATI" and "Intel and Nvidia are the only two stand-alone graphics chip companies in the world." These events are not coreferential.\\ 
                The first event, "AMD acquires ATI," is about the acquisition of ATI Technologies by Advanced Micro Devices. The second event, "Intel and Nvidia are the only two stand-alone graphics chip\\ companies in the world," is about the fact that Intel and Nvidia are the only two companies that make graphics chips that are not part of a larger company.\\
                These two events are not coreferential because they do not share the same participants, locations, times, or actions. The first event involves AMD, ATI, and the acquisition of ATI by AMD.\\ The second event involves Intel, Nvidia, and the fact that they are the only two stand-alone graphics chip companies in the world.\\
                The two events also do not share the same time frame. The first event occurred in 2006, while the second event is a statement about the current state of the graphics chip industry.\\
                Finally, the two events do not share the same action. The first event is about an acquisition, while the second event is about the fact that two companies are the only two in their field.\\
                Therefore, the two events are not coreferential.\\
                The coreferential score is 0.\\
                \textbf{Llama2-7b-chat (Easy to parse \textit{Coreferential results}, True):}\\
                \textit{Coreferential result:} Coreferential\\
                \textit{Coreferential score:} 0.8\\
                Based on the analysis of the two texts, it can be determined that the events mentioned in each text are coreferential to a certain degree. Both events involve the acquisition of ATI Technologies by\\ Advanced Micro Devices, and the impact of this deal on the graphics chip market. The events share several common features, such as the involvement of Intel, Nvidia, and AMD, and the mention\\ of specific dates and locations (July 24 and the computing, mobile communications, and display markets). However, there are also some differences between the two events. For example, the first\\ text focuses more on the reaction of investors and the impact of the deal on AMD's shares, while the second text provides more details on the strategic implications of the deal for AMD and Intel.\\ Based on these observations, it can be concluded that the two events are coreferential to a degree, with a coreferential score of 0.8. This indicates that the events share a significant amount of common\\ information and context, but there are also some differences and nuances that set them apart.

            \end{tabular}\\
    \bottomrule
    \end{tabular}}
    \caption{An illustration of zero-shot CoT prompt for LLM pairwise evaluation and responses of different LLMs. In parentheses after the LLM's name, we record the difficulty in parsing the LLM's predicted label and whether its correctness.}
    \label{tab:zero-shot_CoT}
\end{table*}

\begin{table*}[t]
    \resizebox{\linewidth}{!}{
        \begin{tabular}{l}
            \toprule
            \begin{tabular}[c]{@{}l@{}}
                \textbf{Demonstration:} \\
                We have two texts, each of which contains an event mention that we are interested in, with its head lemma marked with *bold*. Please determine whether these two events are coreferential.\\ Events co-refer only when human participants, non-human participants, locations, times, and actions where events occur co-refer. Your answer should satisfy the following format strictly:\\
                Rationales: Let's think step by step and show the reason here.\\
                Coreferential results: Answer it with only the word `Coreferential' or  `Non-Coreferential'.\\
                Coreferential score: Assign a score between 0 and 1 to measure the degree of co-reference between the two events. A higher numerical value indicates a higher degree of co-reference.\\
                \textit{text1:} Williams-Esther Williams: 'Million Dollar Mermaid' Swimmer And Actress Dies Aged 91. 07 June 2013 Golden girl of screen and pool, Esther Williams, has \textbf{\textit{*died*}} peacefully in her\\ sleep aged 91. It has been confirmed by publicist Harlan Boll that the 1940s Hollywood actress and record-setting swimmer, Esther Williams, died on 6th June in her sleep. Stunning Williams\\ quickly achieved movie success and pin-up status, due to her swimmer's physique being regularly snapped in bathing suits.\\
                \textit{text2:} Esther Williams, Olympic swimmer turned actress and pinup girl, dies at 91 Esther Williams, the swimming champion turned actress who starred in glittering and aquatic Technicolor\\ musicals of the 1940s and 1950s, has died. She was 91. Williams \textbf{\textit{*died*}} early Thursday in her sleep, according to her longtime publicist Harlan Boll. Williams in a bathing suit became a favorite\\ pinup of GI's in World War II, and her popularity continued afterward.\\
                \textit{Rationales:} The participants (Esther Williams) are the same in both events. The location is not explicitly mentioned in either text but can be inferred to be the same. The time of the events aligns\\ closely, with both texts agreeing on the fact that she died at the age of 91 and specifying the same timeframe for her death. The action described in both texts is the death of Esther Williams. The\\ coreference in\\ participants, location, time, and action indicate that the two events are coreferential.\\
                \textit{Coreferential results:} Coreferential\\
                \textit{Coreferential score:} 0.99\\
                \textit{text1:} \textbf{\textit{*Game*}} slips away late as Jaguars fall to Indy Late sacks, too much Manning give Colts comeback win. The Jaguars couldn't make the big play and gave up the big play. That was an all-too-\\familiar scenario Thursday night in a frustrating 31-24 defeat to the Indianapolis Colts.\\
                \textit{text2:} Chiefs lose 20-13 to a fourth quarter Indianapolis Colt stampede Sunday 23 Dec 2012 The Kansas City Chiefs lost possibly the best game they’ve played this season. They lost to a fourth\\ quarter Colt \textbf{\textit{*stampede*}} 20-13. This time it was coach Romeo Crennel’s fault. There they were on the Colts 27-yard line with inches to go for a first down and Crennel decides to go for it. He\\ should have played it safe with a field goal.\\
                \textit{Rationales:} The human participants (teams and key individuals) are different in each event. In the first text, the event is a game between the Jaguars and the Colts, where the Jaguars lose due to\\ late sacks and Peyton Manning's performance. In the second text, the event is a game between the Chiefs and the Colts, where the Chiefs lose due to a decision made by their coach, Romeo Crennel.\\ the locations and times of these events are also different, with one game being on a Thursday night and the other on a Sunday. The actions described (losing a game) are similar but occur in different\\ contexts and for different reasons. These differences indicate that the events are not coreferential.\\
                \textit{Coreferential results:} Non-Coreferential\\
                \textit{Coreferential score:} 0.01\\
                \textit{text1:} A bank in Athens was hit with a fire-bomb during Greek riots, causing the death of at least three people A bank was set on fire by \textbf{\textit{*furious*}} protesters in Athens, resulting in at least three\\ deaths, as they engaged in pitched battles with police over the brutal spending cuts aimed at avoiding national bankruptcy. The Greek capital was plunged into turmoil as stores and businesses were\\ targeted with petrol bombs\\ by hooded youths. The bank and two government buildings caught fire, and rioting spread to Thessaloniki, where police reported three fatalities from the fire-bomb attack.\\
                \textit{text2:} Three dead after fire-bomb hits bank during Greek riots A FIRE-BOMB attack on a bank in Athens left at least three people dead as police fought pitched battles with striking protesters\\ \textbf{\textit{*furious*}} at brutal spending cuts designed to avoid national bankruptcy. Hooded youths hurled petrol bombs at stores and businesses in the Greek capital, setting the bank and two government\\ buildings ablaze. Protesters also tried to storm the Greek Parliament and the rioting spread to Greece's second city, Thessaloniki. Police said at least three people, a man and two women, were\\ killed in the fire-bomb attack.\\
                \textit{Rationales:} The human participants (protesters) are consistent across both events. The location is the same, specified as Athens in both texts. The time of the events is not explicitly mentioned\\ but can be inferred to be the same, as both texts refer to the context of Greek riots and spending cuts. The action is the same, which describes the protesters' anger. The coreference in participants,\\ location, time, and action indicate that the two events are coreferential.\\
                \textit{Coreferential results:} Coreferential\\
                \textit{Coreferential score:} 0.89\\
                \textit{text1:} Lindsay Lohan \textbf{\textit{*Checks Into*}} Betty Ford Rehab–Where She `Knows Everyone' UPDATED 05/03/2013 at 11:00 AM EDT Originally published 05/03/2013 at 10:20 AM EDT Lindsay\\ Lohan checked into the Betty Ford Clinic in Rancho Mirage, Calif., on Thursday night, for what is to be a three-month stay, her rep confirms to PEOPLE.\\
                \textit{text2:} Lindsay Lohan \textbf{\textit{*Checks Out*}} Of Betty Ford Treatment Center And Into Malibu Rehab Posted on Thu Jun 13, 2013 08:45 AM PDT Lindsay Lohan has checked out of the Betty Ford\\ Treatment Center in Palm Springs after completing a 30-day program, but she's not in the clear yet! The Mean Girls star just checked-into the Cliffside Rehab Center in Malibu, where she will\\ receive extended treatment, TMZ reports.\\
                \textit{Rationales:} The participants (Lindsay Lohan) and the initial location (Betty Ford Center) are the same, but the actions (checking in vs. checking out) and the times of these actions (on a Thursday\\ night, though the exact date is not specified vs. on Thursday, June 13, 2013) are different. These differences indicate that the events are not coreferential.\\
                \textit{Coreferential results:} Non-Coreferential\\
                \textit{Coreferential score:} 0.2\\
                \textbf{Input:}\\
                \textit{text1:} Maybe that's why Intel and Nvidia (NVDA), the companies' two biggest rivals, found a lot to smile about on July 24, the day the deal was announced. ``I thought it was just impossible\\ to get a gift like this,'' crowed Nvidia CEO Jen-Hsun Huang, in an interview with BusinessWeek.com. ATI is ``basically throwing in the towel, leaving us as the only stand-alone (graphics\\ chip) company in the world.'' Intel, the world's largest graphics chipmaker, declined to comment specifically on the \textbf{\textit{*deal*}}. But company execs have noted Intel continues to improve its own\\ graphics chipsets and that it expects to deliver platforms of products, like Centrino, that AMD can not easily match. Investors weren't too sure how happy the AMD and ATI family will turn\\ out to be, either. AMD shares fell almost 5\% on July 24, to \$17.40, extending recent declines fueled by concern AMD will be the biggest loser amid an escalating PC industry price war. \\
                \textit{text2:} Addressing a deficiency of its own and further upping the competitive tension between it and chief rival Intel (Nasdaq: INTC-news-people), Advanced Micro Devices (nyse: AMD-news\\-people) intends to acquire graphics chipmaker ATI Technologies (nasdaq: ATYT-news-people ) in a \$5.4 billion deal. Is Intel below \$19 too cheap to last or still above what the chipmaker's\\ worth? The question for many is where that leaves Nvidia Corp. (nasdaq: NVDA - news - people), now the sole independent graphics chip maker addressing each of the computing, mobile\\  communications and display markets. We think not too much worse off than before the \textbf{\textit{*deal*}} and believe shares in the company remain a compelling way to play growth in gaming- and media-\\ oriented computing. Our view of the Advanced Micro Devices/ATI Technologies deal is that it's the last major move AMD needed to make to match Intel's breadth in the processor market (now\\ that Intel has shed its communications processor business). Intel has had for some time the ability to integrate graphics capabilities into its offerings for PC makers, resources for which AMD\\ has had to rely on other folks, including both ATI and Nvidia. The AMD/ATI combination now trumps Intel's strategy by adding the capacity to package higher-end chips for gaming and\\ complex graphic manipulation\\
                \textit{Rationales:} \\
                \textit{Coreferential results:}\\
                \textit{Coreferential score:}  \\
                \textbf{GPT-4 (Easy to parse \textit{Coreferential results}, True)}:\\
                \textit{Rationales:} The participants (Intel, Nvidia, AMD, ATI) are the same in both events. The location is not explicitly mentioned in either text but can be inferred to be the same. The time of the events\\ is not explicitly mentioned but can be inferred to be the same, as both texts refer to the context of the deal between AMD and ATI. The action described in both texts is the deal between AMD and\\ ATI. The coreference in participants, location, time, and action indicate that the two events are coreferential.\\
                \textit{Coreferential results:} Coreferential\\
                \textit{Coreferential score:} 0.95\\
                \textbf{GPT-3.5-turbo (Easy to parse \textit{Coreferential results}, True)}:\\
                \textit{Rationales:} The participants (Intel, Nvidia, AMD, ATI) are consistent across both events. The location is not explicitly mentioned in either text. The times of the events align closely, with both texts\\ referring to the same day (July 24) and the announcement of a deal. The actions described in both texts are related to the deal between AMD and ATI. The coreference in participants, time, and action\\ indicate that the two events are coreferential.\\
                \textit{Coreferential results:} Coreferential\\
                \textit{Coreferential score:} 0.95\\
                
                \textbf{Gemini-Pro (Easy to parse \textit{Coreferential results}, True)}:\\
                \textit{Rationales:} The participants (Intel, AMD, and ATI Technologies) are the same in both events. The location is not explicitly mentioned in either text but can be inferred to be the same, as both texts discuss\\ the companies' positions in the graphics chip market. The time of the events is also not explicitly mentioned but can be inferred to be the same, as both texts refer to the same deal between AMD and ATI\\ Technologies. The action described in both texts is the deal between AMD and ATI Technologies. The coreference in participants, location, time, and action indicate that the two events are coreferential.\\
                \textit{Coreferential results:} Coreferential\\
                \textit{Coreferential score:} 0.95\\
                
                \textbf{Chat-Bison-001 (Easy to parse \textit{Coreferential results}, True)}:\\
                The two events are coreferential. The participants (Intel, Nvidia, AMD, and ATI) are the same in both events. The location is not explicitly mentioned in either text but can be inferred to be the same, as both\\ texts refer to the context of the graphics chip market. The time of the events is also the same, as both texts refer to the July 24, 2006 announcement of the AMD-ATI deal. The action described in both texts\\ is the acquisition of ATI by AMD. The coreference in participants, location, time, and action indicate that the two events are coreferential.\\
                \textit{Coreferential results:} Coreferential\\
                \textit{Coreferential score:} 0.99\\
                \textbf{Llama2-7b-chat (Easy to parse \textit{Coreferential results}, True):}\\
                Sure, I'd be happy to help you with that! Here are the rationales for each event, along with the coreferential result and coreferential score:\\
                1. Input texts:\\
                \textit{text1:} Maybe that's why Intel and Nvidia (NVDA), the companies' two biggest rivals, found a lot to smile about on July 24, the day the deal was announced...AMD shares fell almost 5\% on July 24, to \$17.40,\\ extending recent declines fueled by concern AMD will be the biggest loser amid an escalating PC industry price war.\\
                \textit{Coreferential result:} Coreferential\\
                Coreferential score: 0.99\\
                2. Input texts:\\
                \textit{text1:} Addressing a deficiency of its own and further upping the competitive tension between it and chief rival Intel ... The AMD/ATI combination now trumps Intel's strategy by adding the capacity to package\\ higher-end chips for gaming and complex graphic manipulation\\
                \textit{Coreferential result:} Coreferential\\
                \textit{Coreferential score:} 0.98\\
                Based on the rationales and the coreferential result, I would score the coreferential score between the two events as 0.98, indicating a high degree of co-reference between the two events.\\       
            \end{tabular}\\
    \bottomrule
    \end{tabular}}
    \caption{An illustration of few shots CoT prompt for LLM pairwise evaluation and LLM responses. Llama2-7b-chat provides an implausible response with confused information by repeating text1 and text2 of the input.}
    \label{tab:few-shot_CoT}
\end{table*}

\clearpage
\begin{algorithm*}[t]
    \small
    \caption{Generating Trigger Intervention Ablation (TIA) Data for the original mention pair}
    \label{alg:tia}
    \textbf{Input}: 
    \\Original data $MP$=($S^{\left( 1\right)  }_{i-w}...S^{\left( 1\right)  }_{i}...S^{\left( 1\right)  }_{i+w},S^{\left( 2\right)  }_{j-w}...S^{\left( 2\right)  }_{j}...S^{\left( 2\right)  }_{j+w}$) with label $Y$; Large language model $LLM$; Trigger terms of two mentions ($T^{\left( 1\right)}$,$T^{\left( 2\right)}$).\\
    \textbf{Prompt operators}: Synonyms generator $SYN$; Coref events generator $CE$; Non-coref events generator $NCE$; Paraphraser $PARA$.\\
    \textbf{Output}: Generated dataset: $D_{TIA}$
    \begin{algorithmic}[1] %[1] enables line numbers
    \WHILE{sentence $s$ in $MP$}
        \IF{$Y==coref$}
            \IF {$s==S^{\left( 1\right)  }_{i}$}
                \STATE $S_{gens}=LLM\left( NCE,T^{\left( 1\right)  }, S^{\left( 1\right)  }_{i}\right)$  
            \ELSE 
                \STATE continue
        \ENDIF
        \ELSIF{$Y==not\ coref$}
            \IF {$s==S^{\left( 2\right)  }_{i}$}
                \STATE $S_{gens}=LLM\left( CE,T^{\left( 2\right)  },S^{\left( 2\right)  }_{j}\right)$ 
            \ELSE 
                \STATE continue
            \ENDIF
        \ENDIF
    \ENDWHILE
    \WHILE{sentence $s_{g}$ in $S_{gens}$}
        \IF{$Y==coref$}
            \STATE $\tilde{m}_{1} =(S^{\left( 1\right)  }_{i-w},...,s_{g},...,S^{(1)}_{i+w})$
            \STATE $MP_{cf}=concat\{ \tilde{m}_{1} ,\  (S^{\left( 2\right)  }_{j-w},...,s_{g},...,S^{(2)}_{j+w})\} $
        \ELSIF{$Y==not\ coref$}
            \STATE $pre=LLM\left( PARA,\  (S^{\left( 2\right)  }_{j-w},...,S^{\left( 2\right)  }_{j-1})\right)  $
            \STATE$suf=LLM\left( PARA,\  (S^{\left( 2\right)  }_{j+1},...,S^{\left( 2\right)  }_{j+w})\right)  $
            \STATE $\tilde{m}_{1} =concat\{ pre,s_{g},suf\} $
            \STATE $MP_{TIA}=concat\left\{ \tilde{m}_{1} ,\left( S^{(2)}_{j-w},...,S^{\left( 2\right)  }_{j+w}\right)  \right\}$
        \ENDIF
    \ENDWHILE
    \STATE Add $MP_{TIA}$ to the set $D_{TIA}$
    \STATE \textbf{return} $D_{TIA}$
    \end{algorithmic}
\end{algorithm*}

\begin{algorithm*}[t]
    \small
    \caption{Generating Context Intervention Ablation (CIA) Data for the original mention pair}
    \label{alg:cia}
    \textbf{Input}: 
    \\Original data $MP$=($S^{\left( 1\right)  }_{i-w}...S^{\left( 1\right)  }_{i}...S^{\left( 1\right)  }_{i+w},S^{\left( 2\right)  }_{j-w}...S^{\left( 2\right)  }_{j}...S^{\left( 2\right)  }_{j+w}$) with label $Y$; Large language model $LLM$; Trigger terms of two mentions ($T^{\left( 1\right)}$,$T^{\left( 2\right)}$).\\
    \textbf{Prompt operators}: Synonyms generator $SYN$; Coref events generator $CE$; Non-coref events generator $NCE$; Paraphraser $PARA$.\\
    \textbf{Output}: Generated dataset: $D_{CIA}$
    \begin{algorithmic}[1] %[1] enables line numbers
    \WHILE{sentence $s$ in $MP$}
        \IF{$Y==coref$}
            \IF {$s==S^{\left( 1\right)  }_{i}$}
            \STATE $T^{\left( 1\right)  }_{syns}=LLM\left( SYN,T^{\left( 1\right)  }\right)$
            \STATE $S_{gens}=LLM\left( NCE,T^{\left( 1\right)  }_{syns}, S^{\left( 1\right)  }_{i}\right)$  
            \ELSE 
            \STATE continue
        \ENDIF
        \ELSIF{$Y==not\ coref$}
            \IF {$s==S^{\left( 2\right)  }_{i}$}
            \STATE $T^{\left( 2\right)  }_{syns}=LLM\left( SYN,T^{\left( 2\right)  }\right)$
            \STATE $S_{gens}=LLM\left( CE,T^{\left( 2\right)  }_{syns},S^{\left( 2\right)  }_{j}\right)$ 
            \ELSE 
            \STATE continue
            \ENDIF
        \ENDIF
    \ENDWHILE
    \WHILE{sentence $s_{g}$ in $S_{gens}$}
        \IF{$Y==coref$}
            \STATE $pre^{\left( 1\right)  }=LLM\left( PARA,\  (S^{\left( 1\right)  }_{i-w},...,S^{\left( 1\right)  }_{i-1})\right)  $
            \STATE $suf^{\left( 1\right)  }=LLM\left( PARA,\  (S^{\left( 1\right)  }_{i+1},...,S^{\left( 1\right)  }_{i+w})\right)  $
            \ELSIF{$Y==not\ coref$}
            \STATE $pre^{(1)}=LLM\left( PARA,\  (S^{\left( 2\right)  }_{j-w},...,S^{\left( 2\right)  }_{j-1})\right)$  
            \STATE$suf^{\left( 1\right)  }=LLM\left( PARA,\  (S^{\left( 2\right)  }_{j+1},...,S^{\left( 2\right)  }_{j+w})\right)  $
        \ENDIF
        \STATE $pre^{\left( 2\right)  }=LLM\left( PARA,\  (S^{\left( 2\right)  }_{j-w},...,S^{\left( 2\right)  }_{j-1})\right)$  
        \STATE$suf^{\left( 2\right)  }=LLM\left( PARA,\  (S^{\left( 2\right)  }_{j+1},...,S^{\left( 2\right)  }_{j+w})\right)  $
        \STATE $\tilde{m}_{1} =concat\left\{ pre^{\left( 1\right)  },s_{g},suf^{\left( 1\right)  }\right\}  $
        \STATE $\tilde{m}_{2} =concat\left\{ pre^{\left( 2\right)  },S^{(2)}_{j},suf^{\left( 2\right)  }\right\}  $
        \STATE $MP_{CIA}=concat\{\tilde{m}_{1},\tilde{m}_{2}\} $
    \ENDWHILE
    \STATE Add $MP_{CIA}$ to the set $D_{CIA}$
    \STATE \textbf{return} $D_{CIA}$
    \end{algorithmic}
\end{algorithm*}
        
\begin{algorithm*}[t]
    \small
    \caption{Generating Temporal Commonsense Augmented Data (TAD) for the original mention pair}
    \label{alg:tad}
    \textbf{Input}: 
    \\Original data $MP$=($S^{\left( 1\right)  }_{i-w}...S^{\left( 1\right)  }_{i}...S^{\left( 1\right)  }_{i+w},S^{\left( 2\right)  }_{j-w}...S^{\left( 2\right)  }_{j}...S^{\left( 2\right)  }_{j+w}$) with label $Y$; Large language model $LLM$; Trigger terms of two mentions ($T^{\left( 1\right)}$,$T^{\left( 2\right)}$).\\
    \textbf{Prompt operators}: Synonyms generator $SYN$; Coref events generator $CE$; Non-coref events generator $NCE$; Temporal commonsense generator $TC$.\\
    \textbf{Output}: Generated dataset: $D_{TAD}$
    \begin{algorithmic}[1] %[1] enables line numbers
    \WHILE{sentence $s$ in $MP$}
    \IF{$Y==coref$}
        \IF {$s==S^{\left( 1\right)  }_{i}$}
        \STATE $T^{\left( 1\right)  }_{syns}=LLM\left( SYN,T^{\left( 1\right)  }\right)$
        \STATE $S_{gens}=LLM\left( NCE,T^{\left( 1\right)  }_{syns}, S^{\left( 1\right)  }_{i}\right)$  
        \ELSE 
        \STATE continue
        \ENDIF
    \ELSIF{$Y==not\ coref$}
        \IF {$s==S^{\left( 2\right)  }_{i}$}
        \STATE $T^{\left( 2\right)  }_{syns}=LLM\left( SYN,T^{\left( 2\right)  }\right)$
        \STATE $S_{gens}=LLM\left( CE,T^{\left( 2\right)  }_{syns},S^{\left( 2\right)  }_{j}\right)$ 
        \ELSE 
        \STATE continue
        \ENDIF
    \ENDIF
    \ENDWHILE
    \WHILE{sentence $s_{g}$ in $S_{gens}$}
        \IF{$Y==coref$}
            \STATE $pre^{\left( 1\right)  },suf^{(1)}=LLM\left( TC,\  T^{\left( 1\right)  }_{syns},s_{g}\right)$
        \ELSIF{$Y==not\ coref$}
            \STATE $pre^{\left( 1\right)  },suf^{(1)}=LLM\left( TC,\  T^{\left( 2\right)  }_{syns},s_{g}\right)$
        \ENDIF
        \STATE $pre^{\left( 2\right)  },suf^{(2)}=LLM\left( TC,\  T^{(2)},S^{\left( 2\right)  }_{j}\right)$  
        \STATE $\tilde{m}_{1} =concat\left\{ pre^{\left( 1\right)  },s_{g},suf^{\left( 1\right)  }\right\}  $
        \STATE $\tilde{m}_{2} =concat\left\{ pre^{\left( 2\right)  },S^{(2)}_{j},suf^{\left( 2\right)  }\right\}  $
        \STATE $MP_{TAD}=concat\{\tilde{m}_{1},\tilde{m}_{2}\} $
    \ENDWHILE
    \STATE Add $MP_{TAD}$ to the set $D_{TAD}$
    \STATE \textbf{return} $D_{TAD}$
    \end{algorithmic}
\end{algorithm*}
\end{document}